\documentclass{article}

\usepackage{arxiv}

\usepackage[utf8]{inputenc} 
\usepackage[T1]{fontenc}    
\usepackage{hyperref}       
\usepackage{url}            
\usepackage{booktabs}       
\usepackage{amsfonts}       
\usepackage{nicefrac}       
\usepackage{microtype}      
\usepackage{lipsum}		
\usepackage{graphicx}
\usepackage{doi}

\usepackage[
  backend=biber,      
  style=numeric,       
  sorting=none,
  maxnames=10,
  minnames=1,
  natbib=true
]{biblatex}
\AtEveryBibitem{%
  \iffieldundef{doi}
    {}
    {%
      \clearfield{url}%
      \clearfield{urldate}%
    }%
}

\DeclareSourcemap{
  \maps[datatype=bibtex]{
    \map{
      \step[fieldset=month,   null]
      \step[fieldset=urldate, null]
    }
  }
}

\usepackage{amsmath}
\usepackage{xcolor}
\usepackage{pifont}          
\usepackage{multirow}
\usepackage{comment}
\usepackage[binary-units]{siunitx}
\DeclareSIUnit\rpm{RPM}
\usepackage{subcaption}  

\PassOptionsToPackage{frozencache,cachedir=.}{minted}

\usepackage{minted}
\usepackage{tikz}
\usetikzlibrary{positioning,arrows.meta,calc}

\usepackage{tcolorbox}
\tcbuselibrary{minted,breakable,xparse,skins}

\usepackage{caption}
\usepackage{newfloat}
\usepackage{listings}
\newenvironment{code}{\captionsetup{type=listing}}{}
\SetupFloatingEnvironment{listing}{name=Listing}

\definecolor{bg}{HTML}{FAFAFA}
\definecolor{orangeAPI}{HTML}{FCDE78}

\DeclareTCBListing{mintedbox}{O{}m!O{}}{%
  breakable,
  listing engine=minted,
  listing only,
  minted language=#2,
  minted style=default,
  minted options={%
    linenos,
    gobble=0,
    breaklines,
    breakanywhere=true,           
    breaksymbolleft=,              
    breaksymbolright=,
    breaksymbolindent=0pt,
    breaksymbolsepleft=0pt,
    breaksymbolsepright=0pt,
    fontsize=\small,
    numbersep=8pt,
    #1},
  boxsep=0pt,
  left=25pt,
  top=3pt,
  bottom=3pt,
  arc=5pt,
  leftrule=0pt,
  rightrule=2pt,
  toprule=2pt,
  bottomrule=2pt,
  colback=bg!70,
  colframe=orangeAPI!70,
  enhanced,
  overlay={%
    \begin{tcbclipinterior}
      \fill[orangeAPI!20!white]
        (frame.south west) rectangle ([xshift=20pt]frame.north west);
    \end{tcbclipinterior}},
  #3}

\newcommand{\eg}{{\em e.g.}}

\newcommand{\etal}{{\em et~al.}}

\title{Agentic Large Language Models for Conceptual Systems Engineering and Design}

\author{%
\href{https://orcid.org/0000-0002-6763-3625}{\includegraphics[scale=0.06]{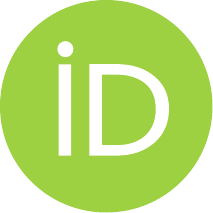}\hspace{1mm}Soheyl Massoudi}\thanks{Corresponding author: \texttt{smassoudi@ehtz.ch}} \\
IDEAL\\
Chair of Artificial Intelligence in Engineering Design \\
ETH Zurich\\
Zurich, Switzerland \\
\texttt{smassoudi@ehtz.ch} \\
\And
\href{https://orcid.org/0000-0003-3819-8895}{\includegraphics[scale=0.06]{orcid.pdf}\hspace{1mm}Mark Fuge} \\
IDEAL\\
Chair of Artificial Intelligence in Engineering Design \\
ETH Zurich\\
Zurich, Switzerland \\
\texttt{mafuge@ethz.ch} \\
}

\renewcommand{\shorttitle}{Agentic Large Language Models for Conceptual Systems Engineering and Design}

\hypersetup{
  pdftitle={Agentic Large Language Models for Conceptual Systems Engineering and Design},
  pdfsubject={cs.AI},
  pdfauthor={Soheyl Massoudi, Mark Fuge},
  pdfkeywords={Large language models, engineering design, autonomous agents, multi-agent systems, design automation, context management, conceptual design, human-AI collaboration},
}

\begin{document}
\maketitle

\begin{abstract}
Early-stage engineering design involves complex, iterative reasoning, yet existing large language model (LLM) workflows struggle to maintain task continuity and generate executable models. We evaluate whether a structured multi-agent system (MAS) can more effectively manage requirements extraction, functional decomposition, and simulator code generation than a simpler two-agent system (2AS). The target application is a solar-powered water filtration system as described in a cahier des charges. 
We introduce the Design-State Graph (DSG), a JSON-serializable representation that bundles requirements, physical embodiments, and Python-based physics models into graph nodes. A nine-role MAS iteratively builds and refines the DSG, while the 2AS collapses the process to a Generator-Reflector loop. Both systems run a total of 60 experiments (2 LLMs - Llama 3.3 70B vs reasoning-distilled DeepSeek R1 70B x 2 agent configurations x 3 temperatures x 5 seeds). We report a JSON validity, requirement coverage, embodiment presence, code compatibility, workflow completion, runtime, and graph size.
Across all runs, both MAS and 2AS maintained perfect JSON integrity and embodiment tagging. Requirement coverage remained minimal (less than 20\%). Code compatibility peaked at 100\% under specific 2AS settings but averaged below 50\% for MAS. Only the reasoning-distilled model reliably flagged workflow completion. Powered by DeepSeek R1 70B, the MAS generated more granular DSGs (average 5-6 nodes) whereas 2AS mode-collapsed.
Structured multi-agent orchestration enhanced design detail. Reasoning-distilled LLM improved completion rates, yet low requirements and fidelity gaps in coding persisted. 
\end{abstract}

\keywords{Large language models \and AI in engineering design \and autonomous agents \and design automation \and local large language models \and human-AI collaboration in design}

\section{Introduction}
Engineering design is an inherently complex, iterative, and multi-objective process, characterized by frequent back-and-forth between problem definition, conceptual exploration, and system integration \cite{wynn_perspectives_2017}. It requires managing evolving requirements, balancing conflicting constraints, and coordinating subsystems to achieve emergent system-level behaviors \cite{salado_contribution_2017, meluso_review_2022}. To navigate these challenges, designers rely on methods such as functional decomposition \cite{she_evaluating_2024}, tacit knowledge formalization \cite{benfell_modeling_2021}, systems structure mining \cite{sexton_organizing_2020}, and the advancement of systems engineering foundations through systems philosophy and methodological synthesis \cite{cook_advances_2021}. Yet, despite these methodological advances, engineering design remains difficult to formalize and automate, with no universally optimal heuristics or models that generalize across domains \cite{watson_engineering_2019}.

Computational tools have helped mitigate parts of the inherent complexity in engineering design. Numerical models, for example, capture the governing physics and interactions between system components \cite{matray_hybrid_2024, lemu_advances_2015}, surrogate models reduce simulation costs for optimization \cite{alizadeh_managing_2020}, and multi-objective optimization frameworks effectively explore trade-offs between competing criteria \cite{sharma_comprehensive_2022, pereira_review_2022}. While these techniques enable efficient analysis of predefined design spaces, they often struggle to accommodate the dynamic, iterative, and ambiguous nature of early-stage design exploration and requirements elicitation \cite{camburn_design_2017}.

Generative artificial intelligence (GenAI) has emerged as a promising alternative in engineering design. Approaches based on Variational Autoencoders (VAEs), Generative Adversarial Networks (GANs), and diffusion models have demonstrated potential in generating complex geometries \cite{chen_beziergan_2021, chen_inverse_2021}, synthesizing innovative topologies \cite{maze_diffusion_2023}, and modeling high-dimensional design spaces \cite{habibi_inverse_2024}, including optimization-aligned and optimization-guided diffusion models that embed performance feedback during generation \cite{giannone_aligning_2023, diniz_optimizing_2024}. These methods enhance the diversity and feasibility of generated designs by integrating performance constraints directly into the generation process. Yet, despite their success, such models remain highly specialized—they require significant domain-specific tuning, focus on narrow tasks, and lack the ability to orchestrate the entire design process from requirements elicitation to final realization \cite{regenwetter_deep_2022}.

Large language models (LLMs) have begun to expand these capabilities by facilitating natural language interaction, requirements extraction, and conceptual ideation \cite{doris_designqa_2024, duan_conceptvis_2024, ataei_elicitron_2024}. Recent work already applies LLMs to tradespace exploration, requirements-engineering analysis, and diagnostics of system engineering-artifact failure modes \cite{apaza_leveraging_2024,norheim_challenges_2024,topcu_trust_2025}. However, in current practice, LLMs serve as static assistants—called upon to generate, summarize, or rephrase design artifacts—rather than as active agents capable of autonomously managing iterative design workflows, invoking simulations, and reasoning across design stages \cite{chiarello_generative_2024}.

In contrast, LLM agents are beginning to transform adjacent fields through tool integration, memory, reflection \cite{wei_chain--thought_2023, yao_react_2023, shinn_reflexion_2023}, and multi-step planning \cite{sumers_cognitive_2024, xie_human-like_2024}. Agent frameworks have been successfully applied in scientific discovery \cite{lu_ai_2024, gottweis_towards_2025}, chemistry \cite{m_bran_augmenting_2024}, biomedical research \cite{gao_empowering_2024}, and software engineering \cite{qian_communicative_2023}, where they autonomously generate hypotheses, execute experiments, and iterate on solutions in complex, knowledge-intensive domains. These successes suggest that engineering design—an inherently iterative, tool-augmented, and knowledge-driven activity—could similarly benefit from agentic LLM architectures.

To the best of our knowledge, LLM agents specifically for conceptual engineering systems design have not yet been investigated \cite{guo_large_2024, wang_survey_2024, picard_concept_2025}. This absence is particularly notable given the alignment between the capabilities of LLM agents—such as subgoal planning \cite{huang_understanding_2024}, tool use \cite{qin_toolllm_2023}, and memory-augmented reasoning \cite{zhang_survey_2024}—and the core challenges of complex system design \cite{vogel_complex_2018}.




Our goal is to evaluate the effectiveness of an LLM-powered multi-agent system for engineering design, specifically in managing long-context reasoning, structured task navigation, and iterative refinement. To validate our approach, we will conduct an ablation study comparing our multi-agent system to a pair of two LLM agents executing the same design tasks within a simple feedback iterative loop. This study aims to determine whether a structured, orchestrated agentic system leads to superior performance in decomposing a high-level engineering goal into functional components, subsystems, and numerical models.

This research is guided by the following research questions:
\begin{itemize}
    \item RQ1: \emph{How effectively can an LLM-powered multi-agent system perform functional decomposition and generate numerical scripts for simulating multi-part, multi-subsystem engineering designs, compared to a simpler two-agent system?}
    \item RQ2: \emph{How accurately does the multi-agent system translate user-specified requirements into functions and physical embodiments within the Design–State Graph?}
\end{itemize}

To investigate the role of structured multi-agent systems in engineering design, we conduct an ablation study comparing our agentic system with a baseline in which a duo of LLM agents, guided by similar contextual system prompts, attempts to extract functions, subfunctions, subsystems, and numerical models using an iterative while-loop structure. This comparison will evaluate whether a multi-agent approach provides advantages in reasoning, coherence, and structured decision-making over an agentic pair acting in isolation.

The study focuses on assessing the ability of each approach to correctly identify and structure key design elements, generate justified numerical models, and maintain logical consistency across design iterations. Design quality, completeness, and efficiency will be evaluated by measuring the correctness of functional decomposition, the coherence of subsystem selection, and the integration of numerical models. Additionally, we will assess whether the structured agentic system improves information management through a graph-based design representation, potentially reducing redundant iterations and accelerating convergence to a structured design.

This investigation is limited to early-stage engineering design, focusing on requirements extraction, functional decomposition, subsystem identification, and initial numerical modeling. It does not cover later design stages such as detailed component-level optimization, multi-objective optimization, or CAD/generative AI-based shape synthesis. The findings are expected to provide empirical evidence on whether an orchestrated multi-agent workflow leads to better-structured and more efficient design processes compared to single-agent reasoning.

\section{Methods}

\subsection{LLM as an AI agent}

\begin{figure*}[ht]%
\centering
\includegraphics[trim={0 5cm 0 2cm}, width=0.8\linewidth]{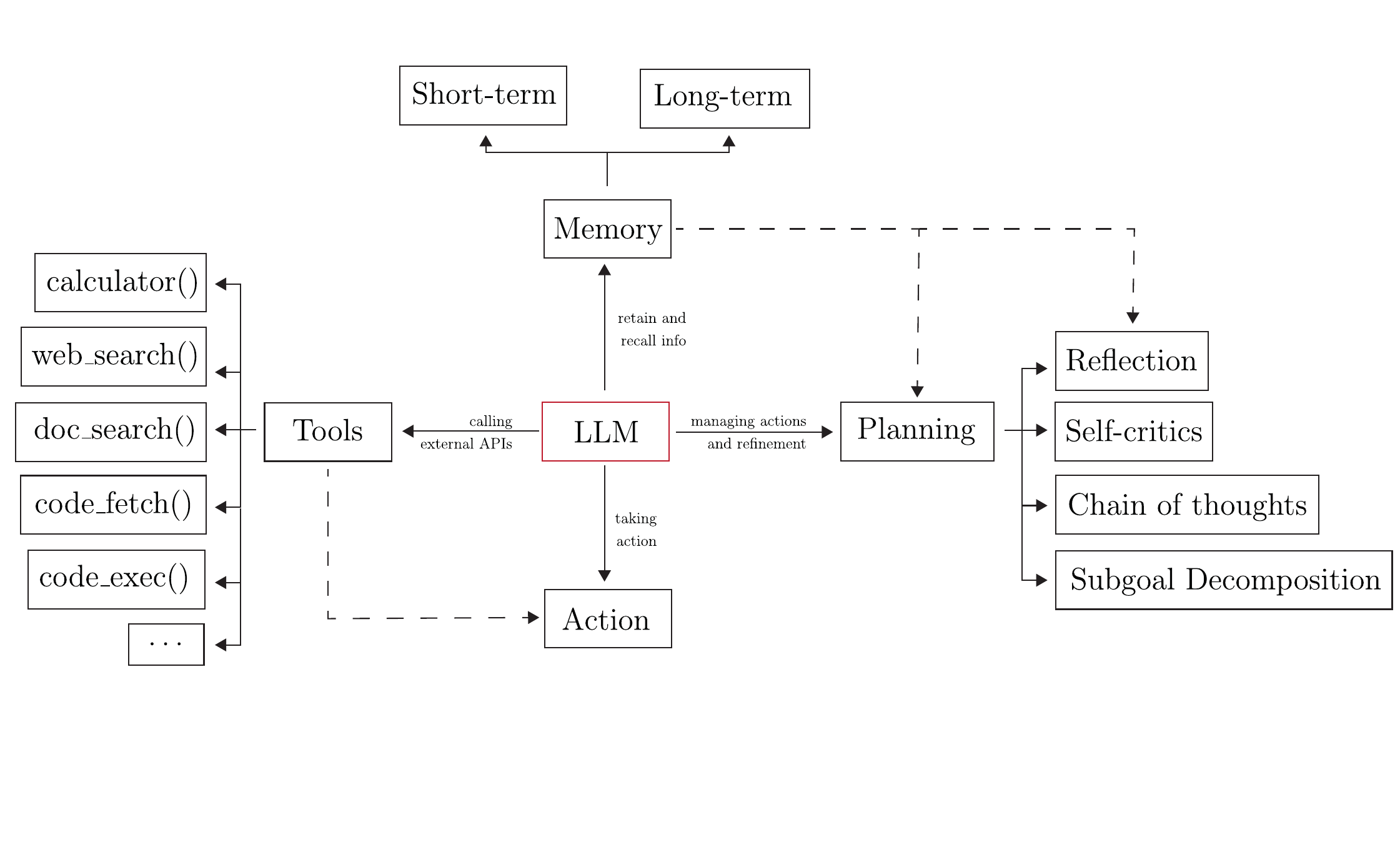}
\caption{The four foundational pillars of an LLM agent—planning, memory, tools, and action. The LLM serves as the core reasoning engine, extended with capabilities for structured decision-making, external tool usage, memory retention, and iterative refinement through planning mechanisms such as reflection and self-criticism. (Concept adapted from \cite{weng2023prompt}.)}
\label{fig:general_four_pillars_LLM_agent}
\end{figure*}

LLMs are widely recognized for their conversational capabilities. However, their deployment as autonomous AI agents represents a shift from reactive text generation to goal-oriented reasoning - one in which the model plans, stores memories, calls external tools and acts upon its environment. In this role, an LLM serves as the core reasoning engine, but its effectiveness hinges on the augmentation of four fundamental capabilities: planning, memory, tool use, and action execution. These four pillars, as presented by \cite{weng2023prompt}, are depicted in Fig.~\ref{fig:general_four_pillars_LLM_agent}.

The integration of these capabilities is made feasible through advancements in function calling via structured JSON outputs, which enable LLMs to interact with external systems programmatically. Server-side libraries further facilitate this process by enforcing structured outputs during chat completions. These advancements align with OpenAI’s introduction of Chat Markup Language (ChatML), which defines a structured sequence of messages:

\begin{itemize}
    \item System message: Defines high-level constraints, including personality traits (\eg, \texttt{You are a helpful assistant.}), domain specialization (\eg, \texttt{You focus on Python, C++, and Java.}), and output format enforcement (\eg, \texttt{Respond using the specified JSON structure.}).
    \item User message: Represents the input from an external entity, typically textual but potentially including images, audio, or video. In this study, we focus solely on text-based interactions.
    \item Assistant message: The LLM-generated response, which may include direct text replies or structured tool calls.
    \item Tool message: Used to transmit the results of a tool execution back to the LLM, allowing it to incorporate external data into its reasoning.
\end{itemize}

A typical ChatML payload is represented as follows:
\begin{code}
\begin{mintedbox}{json}
[
  { "role": "system",
    "content": "You are a helpful assistant" },
  { "role": "user",
    "content": "Explain functional decomposition." }
]
\end{mintedbox}
\captionof{listing}{Example ChatML payload.}
\end{code}

Tool use is a critical enabler of LLM-based agents, transforming them from static information processors into dynamic problem-solvers. Structured JSON-based outputs allow LLMs to execute external API calls, retrieve real-time data from the internet, or interact with databases. This extends their functionality beyond knowledge retrieval to execution-driven problem-solving, including, but not limited to:
\begin{itemize}
    \item Information retrieval: Querying search engines, Wikipedia, or ArXiv to augment responses with up-to-date knowledge.
    \item Code execution: Using environments such as Python REPL to validate hypotheses and generate executable algorithms.
    \item Retrieval-Augmented Generation (RAG): Leveraging vector databases for personalized or domain-specific knowledge retrieval, particularly useful for proprietary or confidential datasets.
\end{itemize}

By incorporating these capabilities, an LLM moves beyond one-shot conversation and can iteratively solve problems - calling external functions, inspecting the results, saving them in memory, and refining its next step. Whenever a tool is invoked, the agent performs an action, can evaluates its outcome, and adjust its plan accordingly. This iterative feedback loop is foundational to intelligent, goal-directed AI systems, aligning with modern cognitive architectures for language agents \cite{sumers_cognitive_2024}.

\subsection{Graph-Based Design Representation}
\label{subsec:dsg}

Early-stage design in our framework is organized around a \emph{Design--State Graph (DSG)}: a typed, directed multigraph that a) captures functional structure and physical realizations, b) carries requirement traceability, and c) bundles executable physics scripts for simulation.

\paragraph{Design--State Graph (DSG).}
We denote the DSG by \(G=(V,E)\), where vertices \(v\in V\) are \emph{design nodes} and directed edges \((v_i,v_j)\in E\) encode flows or dependencies (mass, energy, information). Each design node stores a structured payload that makes the graph self-describing and executable. Figure~\ref{fig:dsg-hierarchy} shows the hierarchy and cardinalities; Table~\ref{tab:dsg-schema} summarizes the schema.

\begin{itemize}
  \item \textbf{DesignState} (1 per DSG): the container snapshot holding
  \texttt{nodes:\{id$\to$payload\}} and a flat \texttt{edges:[(src,dst)]} list, plus a
  \texttt{workflow\_complete} flag for orchestration.
  \item \textbf{DesignNode} (1+ per DesignState): the atomic unit. Besides identifiers
  (\texttt{node\_id}, \texttt{node\_kind}, \texttt{name}), it carries a narrative
  \texttt{description}, maturity/tags, requirement links \texttt{linked\_reqs}, and aggregates
  \emph{exactly one} \textbf{Embodiment} and \emph{zero or more} \textbf{PhysicsModel}s.
  \item \textbf{Embodiment} (exactly 1 per DesignNode): the physical realization of the
  sub-function (\eg\ principle keyword, design parameters with units, coarse cost/mass,
  and a status flag).
  \item \textbf{PhysicsModel} (0+ per DesignNode): an analytical/empirical model with
  human-readable equations, executable \texttt{python\_code}, assumptions, and a status.
\end{itemize}

\paragraph{Edges and traceability.}
Edges are stored as an ordered list of source--target pairs (Fig.~\ref{fig:dsg-hierarchy}, ``DSG edge''), making connectivity explicit and serialization-friendly. Traceability from nodes to requirements is kept in \texttt{linked\_reqs} (right-hand callouts in Fig.~\ref{fig:dsg-hierarchy}). Containment relations (dashed arrows in the figure) clarify the schema-level cardinalities summarized in Table~\ref{tab:dsg-schema}.

\paragraph{Serialization and tooling.}
All payloads are JSON-serializable (lists, dicts, and primitive types). This enables:
(i)~\emph{tool interoperability} (agents read/write well-typed fields; code runners execute \texttt{python\_code} blocks),
and (ii)~\emph{fine-grained versioning} (each agent action mutates the \textbf{DesignState} in memory; periodic snapshots to disk create immutable records of the design evolution).

\paragraph{Design intent.}
The DSG enforces a clean separation of concerns: \emph{function $\rightarrow$ embodiment $\rightarrow$ model}. Agents can thus (a) decompose functions into nodes, (b) choose and revise embodiments without breaking interfaces, and (c) upgrade physics fidelity node-by-node while preserving system connectivity.


\begin{figure}[htbp]
\centering
\begin{tikzpicture}[
  font=\small,
  box/.style={draw, rounded corners, align=center, inner sep=3pt, minimum width=30mm},
  smallbox/.style={draw, rounded corners, align=center, inner sep=2pt, minimum width=28mm},
  edgeA/.style={-Latex, line width=0.5pt},              
  contain/.style={-Latex, dashed, line width=0.5pt},    
  lbl/.style={fill=white, inner sep=1.5pt}
]

\node[box, thick, minimum height=14mm] (ds) {%
  \begin{tabular}{@{}l@{}}
  \textbf{DesignState}\\
  \texttt{nodes:\{id$\to$payload\}}\\
  \texttt{edges:[(src,dst)]}\\
  \texttt{workflow\_complete}
  \end{tabular}
};

\node[box, below left=18mm and 2mm of ds] (dn1) {%
  \textbf{DesignNode 1}\\
  \footnotesize node\_id, name, node\_kind\\[-1pt]
  \footnotesize description, maturity, tags
};
\node[box, below right=18mm and 2mm of ds] (dn2) {%
  \textbf{DesignNode 2}\\
  \footnotesize node\_id, name, node\_kind\\[-1pt]
  \footnotesize description, maturity, tags
};

\node[smallbox, left=8mm of dn1] (reqs1) {%
  \footnotesize \textbf{linked\_reqs}\\ \texttt{\{SR-01, SR-03\}}
};
\node[smallbox, right=8mm of dn2] (reqs2) {%
  \footnotesize \textbf{linked\_reqs}\\ \texttt{\{SR-02\}}
};

\node[smallbox, below=10mm of dn1] (emb1) {%
  \textbf{Embodiment}\\
  \footnotesize principle, parameters\\[-1pt]
  \footnotesize cost\_estimate, mass\_estimate
};
\node[smallbox, below left =8mm and 9mm of emb1] (pm1a) {%
  \textbf{PhysicsModel}\\
  \footnotesize name, equations, code\\[-1pt]
  \footnotesize assumptions, status
};
\node[smallbox, below right=8mm and 9mm of emb1] (pm1b) {%
  \textbf{PhysicsModel}\\
  \footnotesize name, equations, code\\[-1pt]
  \footnotesize assumptions, status
};

\draw[contain] (ds.south) to[bend left=12] node[lbl, midway] {\footnotesize 1..*} (dn1.north);
\draw[contain] (ds.south) to[bend right=12] node[lbl, midway] {\footnotesize 1..*} (dn2.north);

\draw[contain] (dn1.south) -- node[lbl, midway] {\footnotesize exactly 1} (emb1.north);

\coordinate (leftRail)  at ([xshift=-6mm,yshift=2mm]emb1.north west);
\coordinate (rightRail) at ([xshift= 6mm,yshift=2mm]emb1.north east);

\draw[contain] (dn1.south) |- (leftRail)  -| node[lbl, pos=0.60] {\footnotesize 0..*} (pm1a.north);
\draw[contain] (dn1.south) |- (rightRail) -| node[lbl, pos=0.60] {\footnotesize 0..*} (pm1b.north);

\draw[edgeA] (dn1.west) -- (reqs1.east);
\draw[edgeA] (dn2.east) -- (reqs2.west);

\path let \p1 = ($(dn1)!0.5!(dn2)$) in
  node (midpt) at (\x1,\y1) {};
\draw[edgeA] (dn1.east) to[bend left=10] (dn2.west);
\node[lbl] at ($(midpt)+(0,4mm)$) {\footnotesize DSG edge};

\node[align=left, anchor=west, font=\footnotesize] at ($(ds.north east)+(6mm,-1mm)$) {%
\begin{tabular}{@{}l@{}}
\textbf{Notation}\\
Dashed arrows: containment (schema) \\
Solid arrows: DSG edges / associations \\
Edges stored as \texttt{List[List[str]]}
\end{tabular}
};

\end{tikzpicture}
\caption{Hierarchical DSG data model (dashed containment) and a runtime graph edge (solid line). Each \textbf{DesignNode} aggregates exactly one \textbf{Embodiment} and zero or more \textbf{PhysicsModel}s and carries requirement traceability via \texttt{linked\_reqs}. The \textbf{DesignState} stores \texttt{nodes} and a flat \texttt{edges} list as \texttt{List[List[str]]}.}
\label{fig:dsg-hierarchy}
\end{figure}

\begin{table}[htbp]
\centering
\caption{DSG schema summary and cardinalities.}
\label{tab:dsg-schema}
\footnotesize
\resizebox{\columnwidth}{!}{%
\begin{tabular}{@{}l l p{0.62\columnwidth}@{}}
\toprule
\textbf{Component} & \textbf{Cardinality} & \textbf{Key fields (examples)} \\
\midrule
DesignState   & 1 per DSG                & \texttt{nodes:\{id→payload\}}, \texttt{edges:[(src,dst)]}, \texttt{workflow\_complete} \\
DesignNode    & 1+ per DesignState       & \texttt{node\_id}, \texttt{node\_kind}, \texttt{name}, \texttt{linked\_reqs}, \texttt{maturity} \\
Embodiment    & exactly 1 per DesignNode & \texttt{principle}, \texttt{design\_parameters}, \texttt{cost\_estimate}, \texttt{status} \\
PhysicsModel  & 0+ per DesignNode        & \texttt{name}, \texttt{equations}, \texttt{python\_code}, \texttt{assumptions}, \texttt{status} \\
Edge & 0+ per DesignState       & \texttt{src}, \texttt{dst} (stored as \texttt{List[List[str]]}) \\
\bottomrule
\end{tabular}%
}
\end{table}

\subsection{Multi-Agent System}

\begin{figure}[htb]
  \centering
  \begin{subfigure}[t]{\linewidth}
    \centering
    \includegraphics[width=0.7\linewidth]{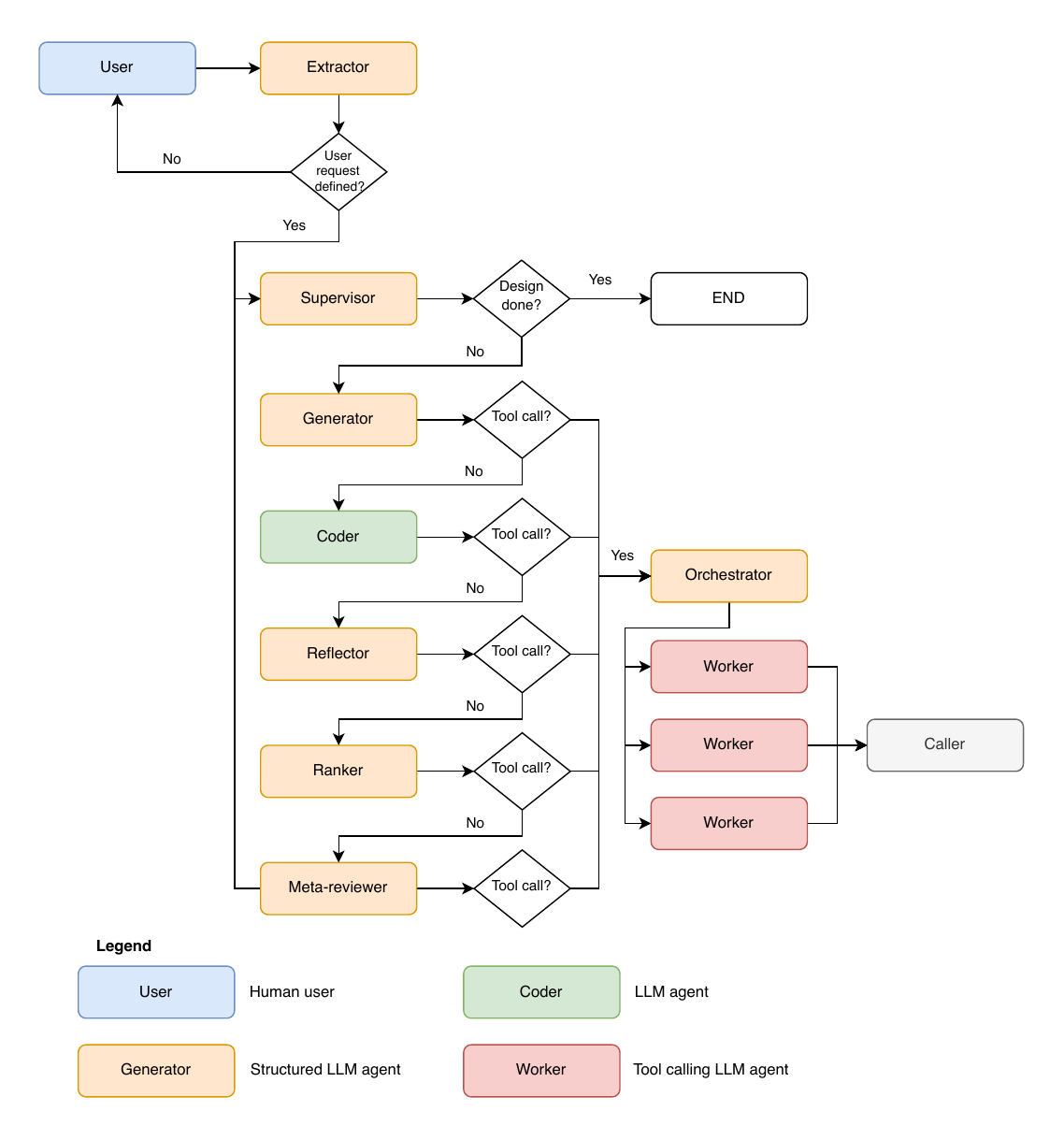}
    \caption{A multi‐agent system (MAS) powered by a LLM for engineering design.}
    \label{fig:MAS}
  \end{subfigure}

  \vspace{1em} 
  \begin{subfigure}[t]{\linewidth}
    \centering
    \includegraphics[width=0.55\linewidth]{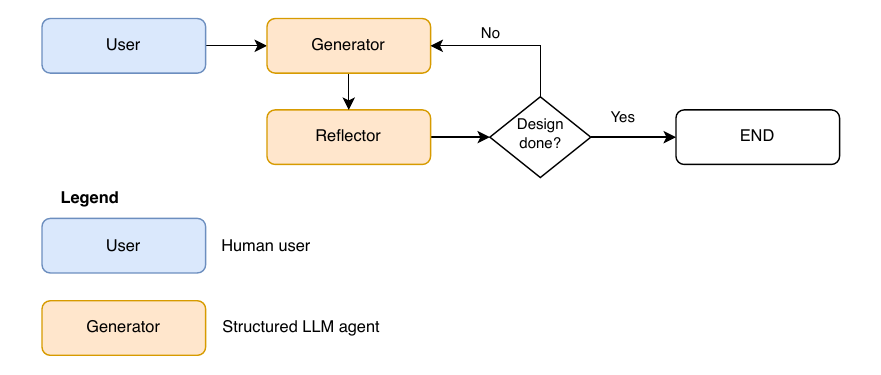}
    \caption{A two‐agent system (2AS) reduced to generation and reflection.}
    \label{fig:2AS}
  \end{subfigure}

  \caption{Comparison of the multi‐agent system and two‐agent system.}
  \label{fig:MASvs2AS}
\end{figure}

A multi-agent system (MAS) is developed to manipulate the DSG through a structured engineering design workflow. The system consists of the following specialized agents as shown in Figure~\ref{fig:MAS}:

\begin{itemize}
    \item \textbf{Extractor Agent} (\ref{subsec:req_prompt}): Interacts with humans to extract and structure project requirements into a formal cahier des charges document using structured JSON output. It iteratively refines requirements until marked as "FINALIZED".
    
    \item \textbf{Supervisor Agent} (\ref{subsec:supervisor_prompt}): Acts as the central coordinator, evaluating design progress and providing step-wise instructions to other agents. It determines whether the current design step is complete or requires iteration, and manages the overall workflow progression.
    
    \item \textbf{Generator Agent} (\ref{subsec:gen_mas_prompt}): Creates multiple DSG proposals (typically three) using LangChain with structured output. Each proposal represents different Pareto-optimal trade-offs in the performance space. The agent can request additional research through the Orchestrator when needed.
    
    \item \textbf{Coder Agent} (\ref{subsec:coder_prompt}): Refines the numerical Python scripts within DSG nodes by processing each physics model in the design graph. It rewrites code to ensure correctness, runnability, and adherence to engineering best practices.
    
    \item \textbf{Reflector Agent} (\ref{subsec:reflection_mas_prompt}): Provides engineering-rigorous critiques of each DSG proposal, evaluating technical soundness, completeness, and compliance with requirements. It may request additional research to strengthen critiques.
    
    \item \textbf{Ranker Agent} (\ref{subsec:ranking_prompt}): Assigns numerical scores (0-10) to each DSG proposal based on alignment with supervisor instructions, compliance with the cahier des charges, and reflection feedback. It may request research to validate rankings.
    
    \item \textbf{Meta-Reviewer Agent} (\ref{subsec:meta_review_prompt}): Selects the best overall DSG proposal from the ranked list and provides detailed instructions for improving the selected solution. It makes final decisions on proposal acceptance, rejection, or iteration needs.
    
    \item \textbf{Orchestrator Agent} (\ref{subsec:orchestrator_prompt}): Manages research tasks by decomposing agent requests into concrete worker tasks involving web searches, ArXiv lookups, or calculations. It coordinates up to three parallel worker tasks.
    
    \item \textbf{Worker Agents} (\ref{subsec:worker_prompt}): Execute specific research tasks using tools including web search and ArXiv search. They provide structured findings and design insights to support agent decision-making. They return the answer to the caller agent.
\end{itemize}

The MAS workflow follows a structured progression: Requirements → Supervisor → Generation → Coder → Reflection → Ranking → Meta-Review, with optional Orchestrator/Worker loops for research needs.

Given the complexity of the MAS workflow, a natural question is whether such complexity is really necessary. To test this, Figure~\ref{fig:2AS} shows a simpler two-agent system (2AS) that we implement as an ablation study compare to the MAS performance. This baseline consists of:

\begin{itemize}
    \item \textbf{Generator Agent} (\ref{subsec:gen_2as_prompt}): Creates DSG proposals using the same structured output approach as the MAS version.
    
    \item \textbf{Reflector Agent} (\ref{subsec:reflection_2as_prompt}): Combines the roles of reflection, ranking, and meta-review from the MAS. It critiques proposals, selects the best one, determines termination criteria, and guides iteration back to generation when needed.
\end{itemize}

Both systems use LangGraph for workflow orchestration with structured state management and checkpointing for experiment reproducibility.

\subsection{Metrics}

To assess our MAS, we focus on its ability to produce runnable numerical scripts for downstream use (\eg, high-fidelity simulation, data generation, optimization). We organize our evaluation along two complementary axes:

\paragraph{Script‐validation hierarchy.}
A truly rigorous assessment would check not only that generated code runs, but also that it is dimensionally correct, meets requirements, and is physically valid. While our present study limits itself to the most basic check—\emph{does the script execute without error?}—we define the full hierarchy below as a roadmap for future work:

\begin{enumerate}
    \item \textbf{Compilation and execution} – Code parses, compiles, and runs without error.
    \item \textbf{Output coherence} – Results carry the correct units and dimensions (\eg, SI).
    \item \textbf{Requirements satisfaction} – Numerical outputs meet the specified design targets.
    \item \textbf{Physics correctness} – Equations and algorithms reflect the governing laws.
    \item \textbf{Fidelity alignment} – Model complexity matches its intended abstraction level.
    \item \textbf{Simulator quality} – Outputs rival those of state-of-the-art engineering solvers.
    \item \textbf{Experimental validation} – Simulations agree (within tolerance) with real-world lab data.
\end{enumerate}

In the experiments that follow we measure Level 1 only. Evaluating Levels 2–7 would require unit-aware prompting, domain-expert review, and vetted benchmark datasets, which are beyond the scope of this study; we leave that assessment to future work. A minimal path forward is feasible: each DSG parameter or model output can be annotated with SI units, enabling lightweight dimensional checks using a unit registry such as \texttt{pint} and symbolic homogeneity tests using \texttt{sympy}. This would elevate validation to Level 2 (dimensional correctness), and can be layered naturally onto the current schema.

\paragraph{Quantitative performance metrics.}
We then measure seven concrete metrics (M1–M7) to compare the MAS with the two-agent baseline:

\begin{itemize}
    \item M1 (JSON validity): Percentage of DSGs that can be parsed (always 1.0 if DSG exists, 0.0 if parsing fails)
    \item M2 (requirements coverage): Percentage of system requirements (SR-01 through SR-10) mentioned in DSG node \texttt{linked\_reqs}, computed via strict exact-match regex for the canonical form $\texttt{SR-XX}$ (variants like \texttt{SR01}/\texttt{SR-001} not counted)
    \item M3 (embodiment presence): Percentage of DSG nodes containing a physical embodiment (embodiment field is not empty)
    \item M4 (code executability): Percentage of Python scripts in DSG nodes that compile and execute successfully (tested via subprocess with --help flag)
    \item M5 (run completion): Number of experimental runs that completed successfully (count of successful runs out of 5)
    \item M6 (wall-clock time): Wall time to completion in seconds (measured from first to last snapshot timestamp)
    \item M7 (graph size): Number of nodes in the final DSG snapshot
\end{itemize}

Together, these metrics capture both the syntactic robustness of our DSG-based representation (M1–M3) and the execution-level quality of the generated numerical code (M4–M7). The above metrics only capture whether the code can run successfully (Level 1), and not whether the generated code is physically accurate or useful (Level 2+).

\subsection{Experiments}

We evaluated every combination of 2 LLM variants, 3 sampling temperatures, and 2 agent configurations across 5 random seeds (0–4):
\[
5\ (\text{seeds})\;\times\;2\ (\text{LLM variants})\;\times\;3\ (\text{temperatures})\;\times\;2\ (\text{agent systems})
\ =\ 60\ \text{experiments}.
\]
Each experiment produces a final DSG, which we evaluate using metrics M1–M7 described above.

All runs target the same engineering case study—a solar-powered water filtration system whose full specifications appear in the cahier des charges in Appendix~\ref{sec:cahier_des_charges}. We varied three independent factors. First, the underlying language model was either the vanilla Llama 3.3 70B (``non-reasoning'') or its ``reasoning''-distilled counterpart, DeepSeek R1 70B. Second, we set the sampling temperature to one of \{0.0, 0.5, 1.0\} to explore the effect of model stochasticity. Third, we compared our full multi-agent system against the simplified two-agent system in which Generation and Reflection alternate in a single loop.

To isolate these effects, we held several elements constant across all runs. We bypassed human-in-the-loop requirement extraction by feeding each agent the finalized cahier des charges and forcing the emission of the `FINALIZED' token. All agents were driven by fixed system prompts (both user-defined and LangChain-internal) to ensure structured JSON outputs. Finally, every experiment executed on the same Nvidia GH200 node under an identical Python environment and vLLM serving framework, guaranteeing that differences in performance arise solely from our chosen independent variables. We also capped the maximum completion tokens at \num{60000}.

The complete LangSmith traces (JSON) and summary table (CSV) for all 60 runs in this study are publicly available on Zenodo \cite{massoudi2025agentic}, and the full implementation codebase is openly available at \url{https://github.com/SoheylM/agentic-eng-design}.

\section{Results}

\subsection{Main Results}
\label{subsec:main-results}

\begin{table}[ht]
  \centering
  \caption{Overall performance (mean\,$\pm$\,std over 5 runs) of each LLM under the multi-agent system (MAS) and two-agent system (2AS) across temperature settings and LLM model. Best values in \textbf{bold}.}
  \label{tab:main-results}
  \resizebox{\columnwidth}{!}{%
  \begin{tabular}{llcccccccc}
    \toprule
    \textbf{LLM} & \textbf{System} & \textbf{Temp} & \textbf{M1 (\%)$\uparrow$} & \textbf{M2 (\%)$\uparrow$} & \textbf{M3 (\%)$\uparrow$} & \textbf{M4 (\%)$\uparrow$} & \textbf{M5 (\#)$\uparrow$} & \textbf{M6 (s)$\downarrow$} & \textbf{M7 (\# N)$\uparrow$} \\
    \midrule
    \multirow{6}{*}{Llama 3.3 70B}
      & \multirow{3}{*}{MAS}
        & 0.0 & \textbf{100\,$\pm$\,0} & 0\,$\pm$\,0 & \textbf{100\,$\pm$\,0} & 40.2\,$\pm$\,3.8 & 0 & 661.7\,$\pm$\,5.2 & 3\,$\pm$\,0 \\
    & & 0.5 & \textbf{100\,$\pm$\,0} & 0\,$\pm$\,0 & \textbf{100\,$\pm$\,0} & 46.8\,$\pm$\,3.7 & 0 & 682\,$\pm$\,242 & 3.75\,$\pm$\,0.96 \\
    & & 1.0 & \textbf{100\,$\pm$\,0} & 0\,$\pm$\,0 & \textbf{100\,$\pm$\,0} & 49\,$\pm$\,28 & 0 & 691\,$\pm$\,153 & 4.6\,$\pm$\,1.3 \\
    \cmidrule{2-10}
      & \multirow{3}{*}{2AS}
        & 0.0 & \textbf{100\,$\pm$\,0} & 0\,$\pm$\,0 & \textbf{100\,$\pm$\,0} & 62.9\,$\pm$\,9.5 & 0 & 1391\,$\pm$\,36 & 5\,$\pm$\,0 \\
    & & 0.5 & \textbf{100\,$\pm$\,0} & 0\,$\pm$\,0 & \textbf{100\,$\pm$\,0} & 94.7\,$\pm$\,3 & 0 & 1192\,$\pm$\,52 & 4\,$\pm$\,0 \\
    & & 1.0 & \textbf{100\,$\pm$\,0} & 0\,$\pm$\,0 & \textbf{100\,$\pm$\,0} & 5\,$\pm$\,12 & 0 & 1097\,$\pm$\,94 & \textbf{6\,$\pm$\,0} \\
    \midrule
    \multirow{6}{*}{DeepSeek R1 70B}
      & \multirow{3}{*}{MAS}
        & 0.0 & \textbf{100\,$\pm$\,0} & \textbf{20\,$\pm$\,0} & \textbf{100\,$\pm$\,0} & 34.8\,$\pm$\,2.1 & 2 & 652\,$\pm$\,452 & 5\,$\pm$\,0 \\
    & & 0.5 & \textbf{100\,$\pm$\,0} & 0\,$\pm$\,0 & \textbf{100\,$\pm$\,0} & 1.3\,$\pm$\,3 & 3 & 225\,$\pm$\,119 & 5.8\,$\pm$\,1.1 \\
    & & 1.0 & \textbf{100\,$\pm$\,0} & 2\,$\pm$\,4.5 & \textbf{100\,$\pm$\,0} & 5\,$\pm$\,12 & 2 & 729\,$\pm$\,619 & \textbf{6\,$\pm$\,1.9} \\
    \cmidrule{2-10}
      & \multirow{3}{*}{2AS}
        & 0.0 & \textbf{100\,$\pm$\,0} & 0\,$\pm$\,0 & \textbf{100\,$\pm$\,0} & \textbf{100\,$\pm$\,0} & \textbf{5} & \textbf{38.6\,$\pm$\,0.68} & 1\,$\pm$\,0 \\
    & & 0.5 & \textbf{100\,$\pm$\,0} & 0\,$\pm$\,0 & \textbf{100\,$\pm$\,0} & 66\,$\pm$\,12 & 3 & 912\,$\pm$\,661 & 2.8\,$\pm$\,1.6 \\
    & & 1.0 & \textbf{100\,$\pm$\,0} & 0.4\,$\pm$\,0.9 & \textbf{100\,$\pm$\,0} & 4\,$\pm$\,8.9 & 0 & 435\,$\pm$\,609 & 1\,$\pm$\,0 \\
  \end{tabular}
  }
  \vspace{0.4em}  
  \caption*{\footnotesize
    \textbf{Metric legend:}
    M1 = JSON validity;
    M2 = requirements coverage;
    M3 = embodiment presence;
    M4 = code executability;
    M5 = run-completion count;
    M6 = wall-clock time (s);
    M7 = graph size (nodes).
  }
\end{table}

\textbf{Key findings.}

Our experiments reveal several consistent patterns in Table~\ref{tab:main-results}. The reasoning-distilled model (DeepSeek R1) not only matches but often outperforms vanilla Llama 3.3 70B across every metric: it is the only model to reliably flag runs as complete (M5 > 0) and, under MAS at T = 1.0, it produces the largest graphs (6 ± 1.9 nodes), suggesting a finer-grained functional decomposition—assuming the additional nodes are distinct rather than duplicates.

The use of the multi-agent architecture drives a clear granularity–runtime trade-off. Focusing on the reasoning LLM, the full MAS produces more granular DSGs (about 6 nodes) but requires roughly \SI{650}{\second} on average, whereas the streamlined 2AS completes in under \SI{40}{\second} with only one node—showing that hierarchical orchestration deepens search at the cost of compute. For the non-reasoning LLM, MAS appears faster than 2AS only because MAS often crashes—either by emitting more than 60 000 tokens or hitting LangGraph’s recursion limit—so the runtime comparison is only meaningful under DeepSeek R1.

Increasing temperature (0.0→1.0) never breaks JSON integrity (M1=100\%) or embodiment tagging (M3=100\%) but does slightly boost node counts (M7) and reduces code compilability (M4), confirming that higher stochasticity favors exploration at the expense of syntactic stability.

Finally, requirement coverage remains minimal—peaking at just 20\% under DeepSeek R1 + MAS + T = 0.0—highlighting that neither agent complexity nor reasoning fine-tuning alone suffices for comprehensive mapping of system requirements into the DSG.

\vspace{0.3em}\noindent\textbf{Answer to RQ1.}  Across 30 runs, MAS with a reasoning LLM achieves up to $6 \pm 1.9$ nodes (M7) and up to $49 \pm 28$\% compilability (M4), demonstrating partial success in decomposition and script generation but falling short of full simulation readiness.

\vspace{0.3em}\noindent\textbf{Answer to RQ2.} Requirement coverage maxes out at 20\% (M2) and embodiment tagging is perfect (M3=100\%), indicating that while the system reliably records physical‐part placeholders, it fails to capture most user requirements.

\subsection{Ablation Study}
\label{subsec:ablation-results}

In our ablation, collapsing to the 2AS trades off decomposition granularity for speed and script validity: for the reasoning-distilled LLM, 2AS outperforms MAS on script executability (M4), run completion (M5), and runtime (M6). Conversely, MAS with the reasoning LLM outperforms 2AS on requirements coverage (M2), graph size (M7), and matches 2AS on JSON validity (M1) and embodiment presence (M3).

\subsection{Qualitative Analysis}
\label{subsec:qualitative}

\begin{figure}[h]
  \centering
  \begin{subfigure}[b]{0.48\linewidth}
    \centering
    \includegraphics[width=\linewidth]{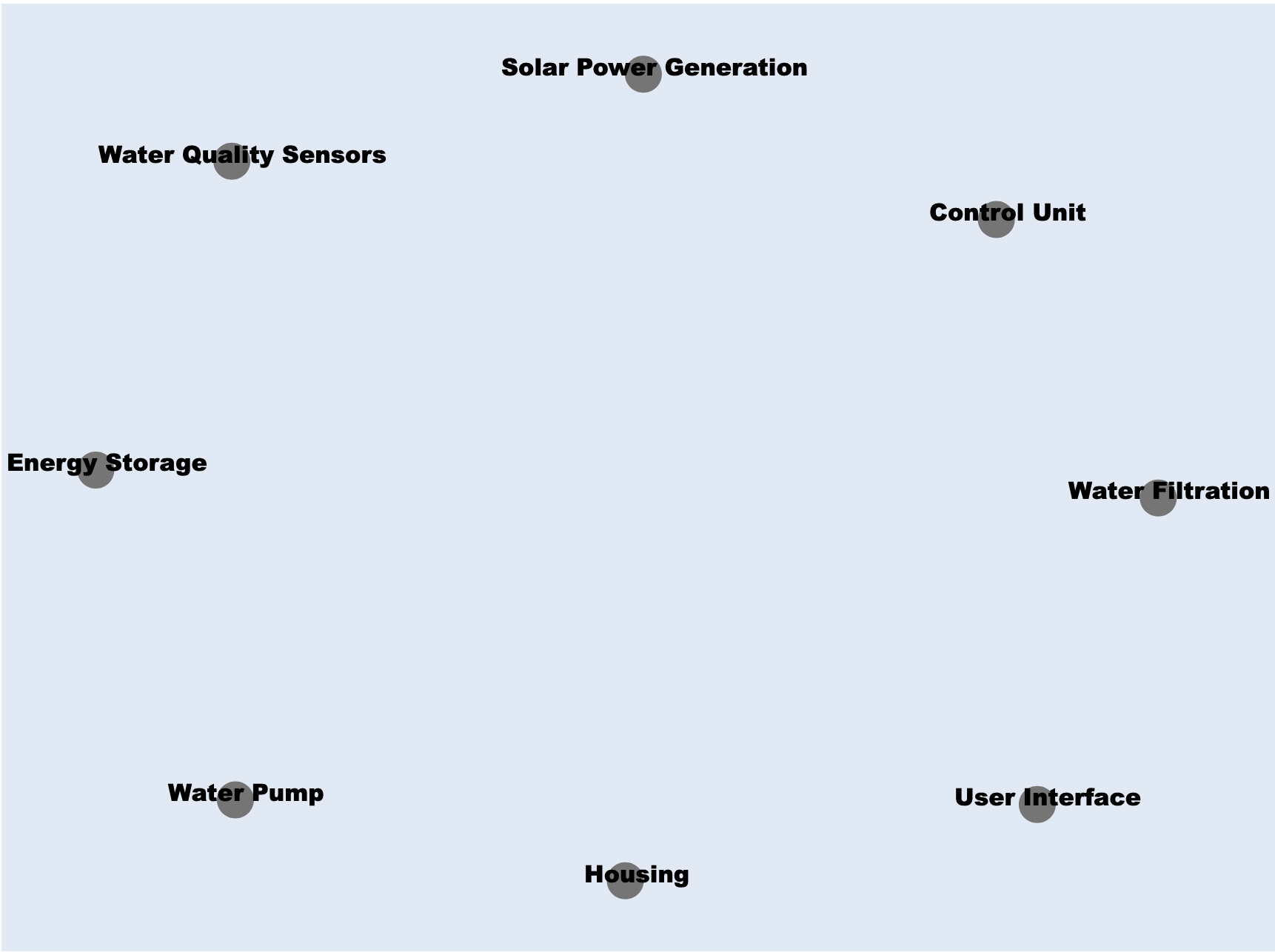}
    \caption{The selected best DSG is produced by a MAS in the fifth run and second iteration for reasoning LLM, temperature 1.0.}
    \label{fig:best_DSG2_MAS}
  \end{subfigure}
  \hfill
  \begin{subfigure}[b]{0.48\linewidth}
    \centering
    \includegraphics[width=\linewidth]{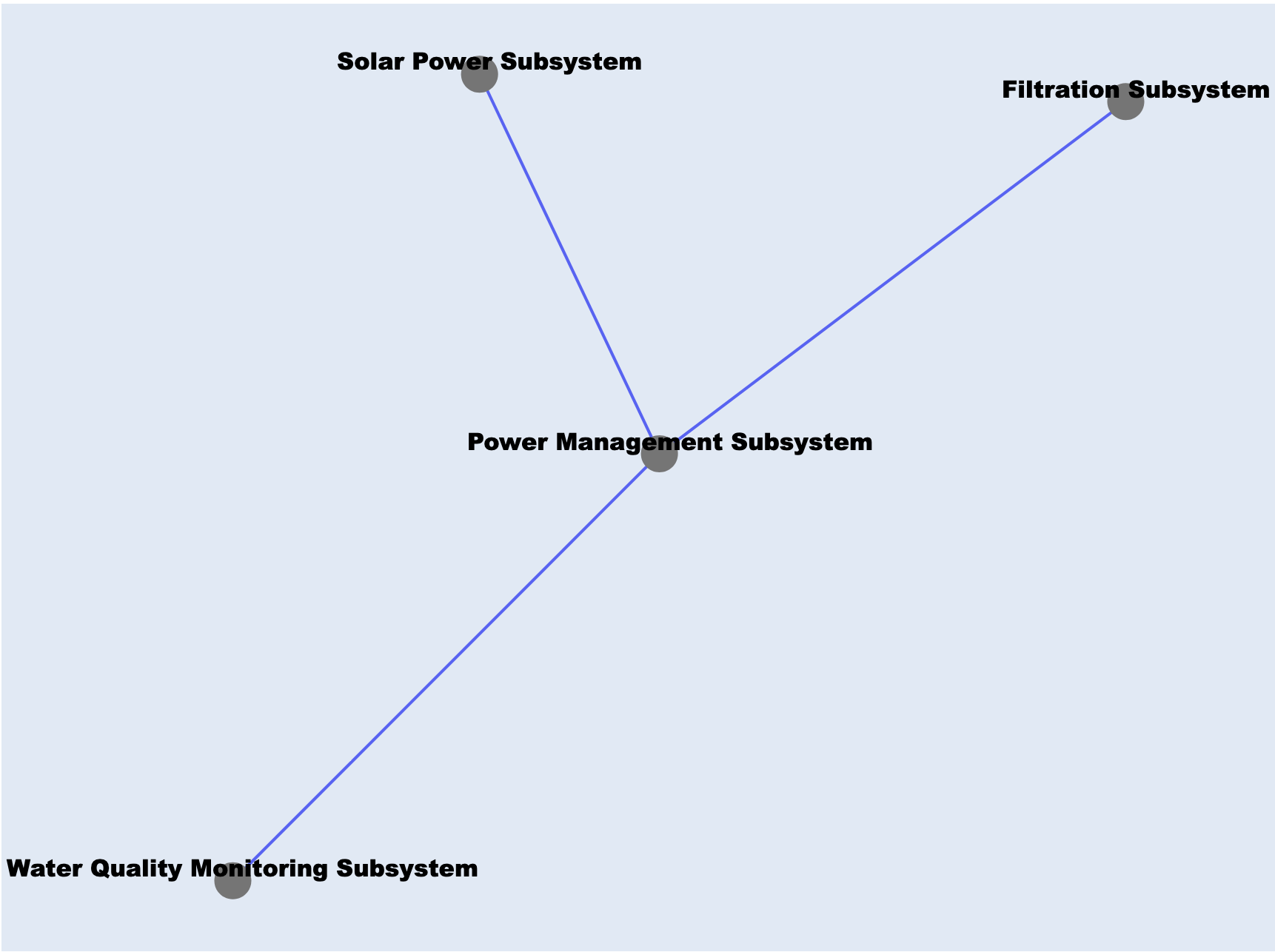}
    \caption{The best DSG produced by 2AS in the first run and 8th iteration of DSG. Reasoning LLM, temperature of 0.5.}
    \label{fig:best_DSG7_2AS}
  \end{subfigure}

  \caption{Comparison of the two best DSGs under MAS and 2AS settings.}
  \label{fig:two_best_DSGs}
\end{figure}

\renewcommand{\arraystretch}{1.25}

\begin{table}[ht]
  \centering
  \caption{Objective comparison of the best MAS and 2AS Design–State Graphs (see code excerpts in Appendix~\ref{sec:app_code_listings}).}
  \label{tab:dsg-comparison}
  \begin{tabular}{@{}p{0.19\linewidth} p{0.37\linewidth} p{0.37\linewidth}@{}}
    \toprule
    \textbf{Criterion} & \textbf{MAS DSG} & \textbf{2AS DSG} \\
    \midrule
    \textbf{Subsystem nodes} &
      8 nodes: power, filtration, pump, storage, control, sensors, UI, housing &
      4 nodes: energy, flow, sensing, battery \\[2pt]

    \textbf{Equation issues} &
      6\,/\,8 scripts dimensionally consistent; two gaps\,: solar model lacks panel area term; pump uses specific gravity instead of density (Listings~\ref{subsec:mas_solar_python_code},~\ref{subsec:mas_pump_python_code}) &
      Uses fixed 300 W m$^{-2}$ irradiance; flow equation omits Darcy relation (Listings~\ref{subsec:2as_energy_python_code},~\ref{subsec:2as_flow_python_code}) \\[2pt]

    \textbf{Script features} &
      Each node script $>$50 lines of code with CLI, logging, file I/O, and unit tests &
      Single 5–15 lines of code function per node; no CLI, logging, I/O, or tests \\[2pt]

    \textbf{Assumption lines} &
      3–6 bullet points per script, enumerating operating limits and simplifications &
      One-line assumption per script, no context \\[2pt]

    \textbf{Represented requirements} &
      Mentions 6 of 10 system requirements in code comments &
      Mentions 1 of 10 requirements \\[2pt]

    \textbf{Simulator readiness} &
      Scripts run out-of-the-box and write CSV/PNG outputs (minor physics fixes still needed) &
      Functions return single values; require wrapper code for any simulation loop \\[2pt]

    \textbf{Modularity} &
      One Python module per node; designed for import or CLI use &
      Stand-alone functions; manual integration needed for workflow \\[2pt]
    \bottomrule
  \end{tabular}
\end{table}

Figure~\ref{fig:two_best_DSGs} and Table~\ref{tab:dsg-comparison} illustrate the stark contrast between the best DSGs generated by the MAS and by the 2AS. In the MAS case (Fig.\ref{fig:best_DSG2_MAS}), the graph spans eight subsystems—solar generation, energy storage, power management, filtration, pump, sensors, user interface, and housing—each accompanied by a Python module complete with a command-line interface, logging, I/O, and unit tests. By contrast, the 2AS output (Fig.\ref{fig:best_DSG7_2AS}) comprises only four nodes. Those nodes implement single‐function stubs (\eg, an energy‐budget function with hard-coded constants) and lack any CLI, I/O support, or testing harness.

The absence of a dedicated Coder agent in the 2AS severely limits code depth: although scripts are present, they offer no algorithmic structure beyond one-line formulas. In the MAS, the Coder agent refines each physics model into a runnable script, but this refinement—looping over every DSG node—adds substantial wall time. Across both systems, edges remain underspecified; mass, energy, and information flows are often declared but not coherently mapped to hierarchical dependencies, indicating room for improved graph‐construction heuristics.

Across all runs, neither MAS nor 2AS fully satisfies the cahier des charges (Appendix~\ref{sec:cahier_des_charges}). The MAS DSG comes closest, covering most physical subsystems and providing detailed simulation code (see Listings~\ref{subsec:mas_solar_python_code}-\ref{subsec:mas_housing_python_code}), but it still omits critical elements: microbiological removal (SR-02), dynamic environmental performance (SR-06), mass and portability estimates linked to SR-07, cost analysis (SR-10), and end-of-life traceability (SR-08). The 2AS DSG is even more limited, entirely excluding energy storage and pump subfunctions and offering only four placeholder scripts (Listings~\ref{subsec:2as_energy_python_code}-\ref{subsec:2as_power_python_code}), with no requirement traceability or I/O support, rendering it incapable of meeting flow (SR-01/SR-03), power (SR-04/SR-05), or usability (SR-09) criteria.

Finally, while Generation agents may invoke external tool calls for research, we observed that the reasoning-distilled MAS seldom requested extra searches—perhaps indicating better internal knowledge—whereas the non-reasoning MAS more frequently triggered worker tasks. This suggests that reasoning fine-tuning not only boosts code completeness but also reduces reliance on external data calls\textemdash either by `competence' or `over-confidence', an avenue worth deeper investigation in future work.

\subsection{Error Analysis}
\label{subsec:error-analysis}

M5 in Table~\ref{tab:main-results} refers to the numbers of runs successfully flagged as completed by the LLM. We manually inspected failed runs. The predominant reason is failure to termination, triggering the max recursion limit (30) of the Langgraph state graph in the MAS or 2AS, as it can be particularly observed with the non-reasoning LLM (Llama 3.3 70B). The second observed failure mode is context overflow, where the LLM emits more than \num{60000} completions tokens in one call.

\section{Discussion}

\label{subsec:insights}

Our study confirms that both Llama 3.3 70B and its reasoning‐distilled variant, DeepSeek R1, achieve perfect JSON validity (M1=100\%), demonstrating the robustness of structured function‐calling fine‐tuning. Yet this fine‐tuning comes at the expense of payload depth: the 2AS prompted for JSON structured output only typically emits minimal Python stubs—scripts that compile (M4) but lack substantive algorithmic detail—whereas the full multi‐agent system (MAS) enriches each node with CLI interfaces, logging, and unit tests at the cost of substantial token exchange.

Under the non‐reasoning LLM, these heavy exchanges can destabilize the model mid‐response. Rather than crashing simply because more agents speak, the shifting context—cycling through varied agent instructions and intermediate outputs—can trigger incoherent, looping outputs (\eg, repeated phrases or nonsensical JSON) that rapidly exhaust the token budget and prevent proper termination (M5). Reasoning distillation (DeepSeek R1) markedly reduces these failures, which seems to align with Meta’s observations on RoPE embeddings and hallucination risks~\cite{grattafiori_llama_2024,xiong_effective_2023}.

Sampling temperature further modulates the exploration–stability tradeoff. Higher temperatures (0.5–1.0) encourage better graph exploration—yielding modest increases in node count (M7)—but also amplify syntax errors, reducing code compilability (M4). In our qualitative review, even the MAS’s strongest DSGs contained fundamental physics errors and unit mismatches in otherwise simple formulas (for example, dimension inconsistencies in solar‐power calculations), underscoring the LLMs’ current limitations in mathematical and domain-grounded reasoning.

Finally, requirement coverage remains low—peaking at 20\% under DeepSeek R1 + MAS + $T{=}0.0$. A manual audit of the experimental runs points to two causes: (i) many nodes omit \texttt{linked\_reqs} altogether (under-linking), and (ii) when links are present, IDs often drift from the canonical CDC format (e.g., \texttt{SR01}, \texttt{SR-001}), which our exact-match evaluator counts as non-matches.

Finally, requirement coverage remains uniformly low—peaking at just 20\% under DeepSeek R1 + MAS + T = 0.0—highlighting that neither agent hierarchy nor reasoning fine‐tuning alone suffices to encode system requirements comprehensively. We found that rigorous regex‐based parsing is essential to avoid both missed matches and false positives in M2 and M4, especially if these metrics become reward signals in subsequent fine-tuning.  

Overall, while agent specialization and reasoning‐distilled models improve graph granularity and completion rates, they fall short of end‐to‐end simulator synthesis. Future work must integrate explicit unit‐aware type systems, automated mathematical verification, and self‐improving agents \cite{zhang_darwin_2025}, capable of refining their own prompts and code to deliver truly physics-correct, production-ready simulation scripts.  
Depending on the application, aligning model behavior with human design judgments should also be considered \cite{keeler_exploring_2025}.

\subsection{Comparison to Prior Work}
\label{subsec:prior}

A variety of recent studies have explored LLM and agentic AI in engineering-design contexts, but none tackle the entire pipeline from requirements to runnable physics models on open-weight architectures. Apaza and Selva apply an LLM dialogue agent to question-answer tradespace exploration, focusing on information retrieval rather than new system synthesis \cite{apaza_leveraging_2024}. Elrefaie \etal build a multi-agent system for aesthetic and aerodynamic car design, yet their agents operate on specialized datasets and produce surrogate evaluations rather than a unified, code-executable graph \cite{elrefaie_ai_2025}.

In concept-generation, actor–critic vison language model (VLM) frameworks, Ghasemi and Moghaddam iteratively refine design sketches via vision-language feedback \cite{ghasemi_vision-language_2025}, and multimodal LLMs have been shown capable of zero-shot 3D CAD program generation \cite{li_llm4cad_2024}. Naghavi Khanghah \etal use graph autoencoders plus an LLM to reconstruct metamaterial topologies, yet their focus remains on node–edge prediction rather than executable simulation code \cite{naghavi_khanghah_reconstruction_2024}. Duan \etal’s ConceptVis integrates LLM-generated design concepts into an interactive knowledge graph to support human-LLM co-ideation, but it does not automate functional decomposition or numerical modeling \cite{duan_conceptvis_2024}.

Alongside these generative methods, structure-mining approaches—such as Sexton and Fuge’s tag-based inference of system ontologies—recover functional relationships from historical records but stop short of driving automated design synthesis or code generation \cite{sexton_organizing_2020}.

By contrast, our framework is, to the best of our knowledge, the first open-weight, multi-agent system that centers on a single, JSON-serializable Design-State Graph, enabling automated requirements decomposition, subsystem mapping, and emission of runnable Python physics models, and their evaluation—bridging the gap between static QA, conceptual ideation, and executable design automation.

\subsection{Limitations}
\label{subsec:limitations}

\begin{itemize}
  \item \textbf{Single design domain.} We evaluated only a solar-powered water filtration system; although our prompts are domain-agnostic and the cahier des charges fully specifies the design scope, transferability to other engineering domains (\eg, turbomachinery, electronics, and so on) remains untested.
  \item \textbf{Partial metric coverage.} This study reports only Level-1 code validity (M4); higher-fidelity physics validation (Levels 2–7) is deferred to future work. A minimal next step is straightforward: each parameter and model output in the DSG can be annotated with SI units and checked using a unit registry such as \texttt{pint}, while libraries like \texttt{sympy} can verify that equations are dimensionally homogeneous. Such lightweight validation would automatically flag the kinds of inconsistencies we observed (e.g., missing area terms in solar models), and would raise the evaluation framework to Level-2 (dimensional correctness). Achieving this would require adaptations to the agent prompts \textemdash most notably extending the Coder agent to emit unit annotations and invoke validation checks, and optionally guiding the Reflector to critique dimensional consistency during review.
  \item \textbf{Compute budget.} Each experiment combination was limited to five seeded runs, reducing statistical power. We exposed only the ArXiv API (excluding web search) and capped the Generator agent at three DSG proposals per cycle, constraining overall design diversity.
  \item \textbf{Model bias.} Llama 3.3 is fine-tuned by Meta on publicly available text, which may introduce domain bias in subsystem suggestions. Its distilled counterpart, DeepSeek R1 70B, inherits—and may amplify—these biases through the distillation process.
  \item \textbf{Prompt engineering bias.} All system prompts were handcrafted and iteratively tuned using closed-source reasoning models (\eg, OpenAI’s o3), creating context-injection biases that strongly shape the DSGs produced by both MAS and 2AS.
\end{itemize}

\subsection{Future Work \& Broader Impact}
Looking ahead, several extensions could substantially strengthen agentic engineering design.  First, we plan to broaden the agents’ toolset—integrating web search, retrieval‐augmented generation, and an interactive Python REPL—using a standardized protocol such as Anthropic’s Model Context Protocol (MCP) to ensure robust tool invocation and maintainable code. Second, powering the Coder agent with an LLM fine‐tuned specifically on simulation code (for example, a Codex‐style model trained on physics scripts) could improve both the depth and correctness of generated modules.  Third, our current executable‐only metric (M4) should be elevated to at least Level 3—verifying dimensional consistency and requirement satisfaction—and ideally to Level 4 for physics correctness; achieving this will likely require a stricter output schema and unit‐aware validation routines. Fourth, fine‐tuning on engineering‐domain corpora offers a promising route to teach models domain expertise, but raises critical questions about dataset bias and the choice of design paradigms (\eg, human‐style vs. purely data‐driven workflows). Finally, extending our framework to multi‐modal LLMs—incorporating vision into the Design‐State Graph—could enrich context and enable agents to reason over diagrams, CAD sketches, or sensor data.

From a broader perspective, fully autonomous conceptual design remains a long‐term goal.  Advances in next‐generation LLMs (\eg, OpenAI o4/o5, future DeepSeek releases, and models from Anthropic, Google, Mistral, and other actors) may close the gap toward end‐to‐end simulation‐ready scripts. However, the risk of deskilling early‐career engineers is real: we therefore advocate for human‐in‐the‐loop safeguards, transparent audit logs of agent decisions, and open review processes to ensure accountability, safety, and equity as these tools mature.

\vspace{0.3em}\noindent\textbf{Take-away.} Our findings suggest that memory-augmented, role-specialized LLM agents chart a promising path toward end-to-end conceptual, multi-part engineering design. Yet the current generation of models cannot deliver fully physics-correct, production-ready simulators for downstream optimization workflows.

\section{Conclusion}

In this paper, we have presented a novel framework for early‐stage conceptual engineering design that harnesses large language models in a structured, multi‐agent architecture centered on a JSON‐serializable Design–State Graph.  By mapping each function to a physical embodiment and an executable Python model, our system enables quantitative evaluation of design workflows under different LLMs (Llama 3.3 70B vs.\ DeepSeek R1 70B), sampling temperatures, and agent configurations (MAS vs.\ 2AS).  Our experiments show that, while reasoning‐distilled models and more developed agent hierarchies deepen exploration and improve termination rates, generated scripts still fall short of physics‐correct, production‐ready simulators, and requirement coverage remains low.  

These results underscore both the promise and the current limitations of LLM‐powered design co‐pilots: structured orchestration and fine‐tuning boost graph granularity and workflow completion, but fundamental gaps in mathematical rigor and domain reasoning persist.  Looking forward, integrating unit‐aware type systems, automated verification modules, and self‐improving prompt‐and‐code loops will be critical to close these gaps.  We believe our DSG‐based multi‐agent paradigm lays important groundwork for AI assistants that can autonomously navigate the earliest, most open‐ended phases of complex, multi‐part engineering design.

\section*{Glossary}

\begin{description}

\item[\textbf{Cahier des Charges (CDC)}]
Structured requirements specification used as the single source of truth for the target design task.

\item[\textbf{Design–State Graph (DSG)}]
Typed, directed multigraph $G=(V,E)$ representing the evolving design.
Nodes (\emph{DesignNodes}) hold structured payloads; edges encode flows or dependencies.

\item[\textbf{DesignState}]
Serialized snapshot of the whole DSG containing a map of node IDs to payloads and a flat edge list.
Also stores a \texttt{workflow\_complete} flag for orchestration.

\item[\textbf{DesignNode}]
Atomic unit of the DSG. Includes identifiers (\texttt{node\_id}, \texttt{node\_kind}, \texttt{name}), description, maturity/tags,
\texttt{linked\_reqs} (traceability), exactly one \textbf{Embodiment}, and zero or more \textbf{PhysicsModel}s.

\item[\textbf{Embodiment}]
Physical realization of a function (e.g., reverse osmosis module); includes principle, design parameters (with units),
coarse cost/mass, and a status flag.

\item[\textbf{PhysicsModel}]
Analytical/empirical model attached to a node (equations, executable \texttt{python\_code}, assumptions, status).
Intended for simulation or quick checks.

\item[\textbf{linked\_reqs}]
Per-node list of requirement IDs (e.g., SR-01) indicating which CDC items the node claims to satisfy.

\item[\textbf{DSG Edge}]
Directed source–destination pair (stored as \texttt{[src, dst]}) representing a flow or dependency between nodes.

\item[\textbf{Requirements Extraction}]
Process of parsing the CDC to identify and structure functional/non-functional ``shall'' statements,
constraints, and verification methods; yields normalized IDs (e.g., SR-01) and optional \texttt{linked\_reqs} seeds.

\item[\textbf{Multi-Agent System (MAS)}]
Nine-role LLM workflow (Extractor, Supervisor, Generator, Coder, Reflector, Ranker, Meta-Review, Orchestrator, Workers)
that iteratively builds and refines the DSG.

\item[\textbf{Two-Agent System (2AS)}]
Ablation baseline using only \emph{Generator} and \emph{Reflector} in a loop.

\item[\textbf{Extractor}]
LLM role that conducts \emph{Requirements Extraction}: interacts (or is provided the CDC) to structure the specification,
normalize requirement IDs, and mark the CDC as \texttt{FINALIZED} for downstream agents.

\item[\textbf{Supervisor}]
Coordinator that issues step-wise instructions (generate, critique, rank, iterate) and decides when the workflow is complete.

\item[\textbf{Generator}]
Produces candidate DSGs (functional decomposition, embodiments, and models) under structured output constraints.

\item[\textbf{Coder}]
Refactors/improves per-node \texttt{python\_code} for runnability and basic quality (CLI, logging, I/O), given the selected DSG.

\item[\textbf{Reflector}]
Provides engineering critiques of candidate DSGs against the CDC and Supervisor instructions; suggests actionable fixes.

\item[\textbf{Ranker}]
Scores candidate DSGs (0–10) using Supervisor goals, CDC coverage, and Reflector feedback.

\item[\textbf{Meta-Review}]
Selects the best DSG proposal and issues improvement instructions to focus subsequent iterations.

\item[\textbf{Orchestrator / Workers}]
Decomposes research requests (e.g., literature lookups, quick calcs) into concrete tasks; Workers execute and return structured findings.

\item[\textbf{ChatML}]
Structured message format for LLM chat (roles: \emph{system}, \emph{user}, \emph{assistant}, \emph{tool}) enabling function/tool calls
and deterministic JSON outputs.

\item[\textbf{Function calling}]
LLM emits a structured JSON ``tool call'' that an external runner executes; results are returned as a \emph{tool} message.

\item[\textbf{LangGraph}]
Graph-based orchestration library used to implement agent workflows, state, and checkpointing.
A ``recursion limit'' prevents non-terminating loops during iteration.

\item[\textbf{Reasoning-distilled model}]
LLM variant tuned for longer-horizon reasoning heuristics (here, DeepSeek R1 70B) compared to a vanilla model (Llama 3.3 70B).

\item[\textbf{Sampling temperature}]
Decoding parameter controlling randomness in token selection; higher values explore more but may reduce syntactic stability.

\item[\textbf{Level-1 executability (M4 scope)}]
Whether generated \texttt{python\_code} runs without error; does \emph{not} verify units, dimensional consistency, or physics correctness.

\item[\textbf{Metrics (M1–M7)}]
\begin{itemize}\itemsep2pt
  \item \textbf{M1} JSON validity (parsable DSG).
  \item \textbf{M2} Requirement coverage (share of CDC IDs mentioned in nodes).
  \item \textbf{M3} Embodiment presence in nodes.
  \item \textbf{M4} Code executability (Level-1).
  \item \textbf{M5} Run completion count.
  \item \textbf{M6} Wall-clock time to completion.
  \item \textbf{M7} Graph size (final node count).
\end{itemize}

\item[\textbf{vLLM}]
High-throughput inference server used to host the underlying LLMs for the experiments.

\end{description}

\printbibliography

@article{alizadeh_managing_2020,
 author = {Alizadeh, Reza and Allen, Janet K. and Mistree, Farrokh},
 doi = {10.1007/s00163-020-00336-7},
 journal = {Research in Engineering Design},
 number = {3},
 pages = {275--298},
 title = {Managing computational complexity using surrogate models: a critical review},
 volume = {31},
 year = {2020}
}

@article{apaza_leveraging_2024,
 author = {Apaza, Gabriel and Selva, Daniel},
 doi = {10.2514/1.A35834},
 journal = {Journal of Spacecraft and Rockets},
 number = {5},
 pages = {1165--1183},
 title = {Leveraging {Large} {Language} {Models} for {Tradespace} {Exploration}},
 volume = {61},
 year = {2024}
}

@inproceedings{ataei_elicitron_2024,
 author = {Ataei, Mohammadmehdi et al.},
 doi = {10.1115/DETC2024-143598},
 publisher = {American Society of Mechanical Engineers Digital Collection},
 title = {Elicitron: {A} {Framework} for {Simulating} {Design} {Requirements} {Elicitation} {Using} {Large} {Language} {Model} {Agents}},
 year = {2024}
}

@article{benfell_modeling_2021,
 author = {Benfell, Adrian},
 doi = {10.1007/s00766-020-00330-4},
 journal = {Requirements Engineering},
 number = {1},
 pages = {25--42},
 title = {Modeling functional requirements using tacit knowledge: a design science research methodology informed approach},
 volume = {26},
 year = {2021}
}

@article{camburn_design_2017,
 author = {Camburn, Bradley et al.},
 doi = {10.1017/dsj.2017.10},
 journal = {Design Science},
 pages = {e13},
 title = {Design prototyping methods: state of the art in strategies, techniques, and guidelines},
 volume = {3},
 year = {2017}
}

@misc{chen_beziergan_2021,
 author = {Chen, Wei and Fuge, Mark},
 doi = {10.48550/arXiv.1808.08871},
 publisher = {arXiv},
 title = {B{ézierGAN}: {Automatic} {Generation} of {Smooth} {Curves} from {Interpretable} {Low}-{Dimensional} {Parameters}},
 year = {2021}
}

@article{chen_inverse_2021,
 author = {Chen, Qiuyi and Wang, Jun and Pope, Phillip and Chen, Wei (Wayne) and Fuge, Mark},
 doi = {10.1115/1.4052846},
 journal = {Journal of Mechanical Design},
 number = {021712},
 title = {Inverse {Design} of {Two}-{Dimensional} {Airfoils} {Using} {Conditional} {Generative} {Models} and {Surrogate} {Log}-{Likelihoods}},
 volume = {144},
 year = {2021}
}

@article{chiarello_generative_2024,
 author = {Chiarello, Filippo and Barandoni, Simone and Škec, Marija Majda and Fantoni, Gualtiero},
 doi = {10.1017/pds.2024.198},
 journal = {Proceedings of the Design Society},
 pages = {1959--1968},
 title = {Generative large language models in engineering design: opportunities and challenges},
 volume = {4},
 year = {2024}
}

@article{cook_advances_2021,
 author = {Cook, Stephen C. and Pratt, Jaci M.},
 doi = {10.1080/14488388.2020.1809845},
 journal = {Australian Journal of Multi-Disciplinary Engineering},
 number = {1},
 pages = {9--22},
 title = {Advances in systems of systems engineering foundations and methodologies},
 volume = {17},
 year = {2021}
}

@article{doris_designqa_2024,
 author = {Doris, Anna C. et al.},
 doi = {10.1115/1.4067333},
 journal = {Journal of Computing and Information Science in Engineering},
 pages = {1--16},
 title = {{DesignQA}: {A} {Multimodal} {Benchmark} for {Evaluating} {Large} {Language} {Models}' {Understanding} of {Engineering} {Documentation}},
 year = {2024}
}

@inproceedings{duan_conceptvis_2024,
 author = {Duan, Runlin et al.},
 doi = {10.1115/DETC2024-146409},
 publisher = {American Society of Mechanical Engineers Digital Collection},
 title = {{ConceptVis}: {Generating} and {Exploring} {Design} {Concepts} for {Early}-{Stage} {Ideation} {Using} {Large} {Language} {Model}},
 year = {2024}
}

@misc{elrefaie_ai_2025,
 author = {Elrefaie, Mohamed and Qian, Janet and Wu, Raina and Chen, Qian and Dai, Angela and Ahmed, Faez},
 doi = {10.48550/arXiv.2503.23315},
 publisher = {arXiv},
 title = {{AI} {Agents} in {Engineering} {Design}: {A} {Multi}-{Agent} {Framework} for {Aesthetic} and {Aerodynamic} {Car} {Design}},
 year = {2025}
}

@article{gao_empowering_2024,
 author = {Gao, Shanghua and Fang, Ada and Huang, Yepeng and Giunchiglia, Valentina and Noori, Ayush and Schwarz, Jonathan Richard and Ektefaie, Yasha and Kondic, Jovana and Zitnik, Marinka},
 doi = {10.1016/j.cell.2024.09.022},
 journal = {Cell},
 number = {22},
 pages = {6125--6151},
 title = {Empowering biomedical discovery with {AI} agents},
 volume = {187},
 year = {2024}
}

@article{ghasemi_vision-language_2025,
 author = {Ghasemi, Parisa and Moghaddam, Mohsen},
 doi = {10.1115/1.4067619},
 journal = {Journal of Mechanical Design},
 pages = {1--28},
 title = {Vision-{Language} {Models} for {Design} {Concept} {Generation}: {An} {Actor}-{Critic} {Framework}},
 year = {2025}
}

@misc{gottweis_towards_2025,
 author = {Gottweis, Juraj and Weng, Wei-Hung and Daryin, Alexander and Tu, Tao and Palepu, Anil and Sirkovic, Petar and Myaskovsky, Artiom and Weissenberger, Felix and Rong, Keran and Tanno, Ryutaro and others},
 doi = {10.48550/arXiv.2502.18864},
 publisher = {arXiv},
 title = {Towards an {AI} co-scientist},
 year = {2025}
}

@misc{grattafiori_llama_2024,
 author = {Grattafiori, Aaron and Dubey, Abhimanyu and Jauhri, Abhinav and Pandey, Abhinav and Kadian, Abhishek and Al-Dahle, Ahmad and Letman, Aiesha and Mathur, Akhil and Schelten, Alan and Vaughan, Alex and others},
 doi = {10.48550/arXiv.2407.21783},
 publisher = {arXiv},
 title = {The {Llama} 3 {Herd} of {Models}},
 year = {2024}
}

@misc{guo_large_2024,
 author = {Guo, Taicheng and Chen, Xiuying and Wang, Yaqi and Chang, Ruidi and Pei, Shichao and Chawla, Nitesh V. and Wiest, Olaf and Zhang, Xiangliang},
 doi = {10.48550/arXiv.2402.01680},
 publisher = {arXiv},
 title = {Large {Language} {Model} based {Multi}-{Agents}: {A} {Survey} of {Progress} and {Challenges}},
 year = {2024}
}

@inproceedings{habibi_inverse_2024,
 author = {Habibi, Milad and Fuge, Mark},
 doi = {10.1115/DETC2024-143607},
 publisher = {American Society of Mechanical Engineers Digital Collection},
 title = {Inverse {Design} {With} {Conditional} {Cascaded} {Diffusion} {Models}},
 year = {2024}
}

@misc{huang_understanding_2024,
 author = {Huang, Xu and Liu, Weiwen and Chen, Xiaolong and Wang, Xingmei and Wang, Hao and Lian, Defu and Wang, Yasheng and Tang, Ruiming and Chen, Enhong},
 doi = {10.48550/arXiv.2402.02716},
 publisher = {arXiv},
 title = {Understanding the planning of {LLM} agents: {A} survey},
 year = {2024}
}

@article{lemu_advances_2015,
 author = {Lemu, Hirpa G.},
 doi = {10.1007/s40436-015-0110-9},
 journal = {Advances in Manufacturing},
 number = {2},
 pages = {130--138},
 title = {Advances in numerical computation based mechanical system design and simulation},
 volume = {3},
 year = {2015}
}

@article{li_llm4cad_2024,
 author = {Li, Xingang and Sun, Yuewan and Sha, Zhenghui},
 doi = {10.1115/1.4067085},
 journal = {Journal of Computing and Information Science in Engineering},
 number = {021005},
 title = {{LLM4CAD}: {Multimodal} {Large} {Language} {Models} for {Three}-{Dimensional} {Computer}-{Aided} {Design} {Generation}},
 volume = {25},
 year = {2024}
}

@misc{lu_ai_2024,
 author = {Lu, Chris and Lu, Cong and Lange, Robert Tjarko and Foerster, Jakob and Clune, Jeff and Ha, David},
 doi = {10.48550/arXiv.2408.06292},
 publisher = {arXiv},
 title = {The {AI} {Scientist}: {Towards} {Fully} {Automated} {Open}-{Ended} {Scientific} {Discovery}},
 year = {2024}
}

@article{m_bran_augmenting_2024,
 author = {M. Bran, Andres and Cox, Sam and Schilter, Oliver and Baldassari, Carlo and White, Andrew D. and Schwaller, Philippe},
 doi = {10.1038/s42256-024-00832-8},
 journal = {Nature Machine Intelligence},
 number = {5},
 pages = {525--535},
 title = {Augmenting large language models with chemistry tools},
 volume = {6},
 year = {2024}
}

@dataset{massoudi2025agentic,
  author       = {Massoudi, Soheyl and Fuge, Mark},
  title        = {Agentic LLM traces for Conceptual Systems Engineering \& Design},
  year         = {2025},
  publisher    = {Zenodo},
  doi          = {10.5281/zenodo.15855569},
  url          = {https://doi.org/10.5281/zenodo.15855569},
  type         = {Dataset}
}

@article{matray_hybrid_2024,
 author = {Matray, Victor et al.},
 doi = {10.1016/j.cma.2024.117243},
 journal = {Computer Methods in Applied Mechanics and Engineering},
 pages = {117243},
 title = {A hybrid numerical methodology coupling reduced order modeling and {Graph} {Neural} {Networks} for non-parametric geometries: {Applications} to structural dynamics problems},
 volume = {430},
 year = {2024}
}

@article{maze_diffusion_2023,
 author = {Mazé, François and Ahmed, Faez},
 doi = {10.1609/aaai.v37i8.26093},
 journal = {Proceedings of the AAAI Conference on Artificial Intelligence},
 number = {8},
 pages = {9108--9116},
 title = {Diffusion {Models} {Beat} {GANs} on {Topology} {Optimization}},
 volume = {37},
 year = {2023}
}

@article{meluso_review_2022,
 author = {Meluso, John et al.},
 doi = {10.1109/TSMC.2022.3163019},
 journal = {IEEE Transactions on Systems, Man, and Cybernetics: Systems},
 number = {12},
 pages = {7679--7691},
 title = {A {Review} and {Framework} for {Modeling} {Complex} {Engineered} {System} {Development} {Processes}},
 volume = {52},
 year = {2022}
}

@article{naghavi_khanghah_reconstruction_2024,
 author = {Naghavi Khanghah, Kiarash and Wang, Zihan and Xu, Hongyi},
 doi = {10.1115/1.4066095},
 journal = {Journal of Computing and Information Science in Engineering},
 number = {021003},
 title = {Reconstruction and {Generation} of {Porous} {Metamaterial} {Units} {Via} {Variational} {Graph} {Autoencoder} and {Large} {Language} {Model}},
 volume = {25},
 year = {2024}
}

@article{norheim_challenges_2024,
 author = {Norheim, Johannes J. and Rebentisch, Eric and Xiao, Dekai and Draeger, Lorenz and Kerbrat, Alain and Weck, Olivier L. de},
 doi = {10.1017/dsj.2024.8},
 journal = {Design Science},
 pages = {e16},
 title = {Challenges in applying large language models to requirements engineering tasks},
 volume = {10},
 year = {2024}
}

@article{pereira_review_2022,
 author = {Pereira, João Luiz Junho et al.},
 doi = {10.1007/s11831-021-09663-x},
 journal = {Archives of Computational Methods in Engineering},
 number = {4},
 pages = {2285--2308},
 title = {A {Review} of {Multi}-objective {Optimization}: {Methods} and {Algorithms} in {Mechanical} {Engineering} {Problems}},
 volume = {29},
 year = {2022}
}

@misc{qian_communicative_2023,
 author = {Qian, Chen and Cong, Xin and Liu, Wei and Yang, Cheng and Chen, Weize and Su, Yusheng and Dang, Yufan and Li, Jiahao and Xu, Juyuan and Li, Dahai and others},
 publisher = {arXiv},
 title = {Communicative {Agents} for {Software} {Development}},
 year = {2023}
}

@misc{qin_toolllm_2023,
 author = {Qin, Yujia and Liang, Shihao and Ye, Yining and Zhu, Kunlun and Yan, Lan and Lu, Yaxi and Lin, Yankai and Cong, Xin and Tang, Xiangru and Qian, Bill and others},
 doi = {10.48550/arXiv.2307.16789},
 publisher = {arXiv},
 title = {{ToolLLM}: {Facilitating} {Large} {Language} {Models} to {Master} 16000+ {Real}-world {APIs}},
 year = {2023}
}

@article{regenwetter_deep_2022,
 author = {Regenwetter, Lyle and Nobari, Amin Heyrani and Ahmed, Faez},
 doi = {10.1115/1.4053859},
 journal = {Journal of Mechanical Design},
 number = {071704},
 title = {Deep {Generative} {Models} in {Engineering} {Design}: {A} {Review}},
 volume = {144},
 year = {2022}
}

@article{salado_contribution_2017,
 author = {Salado, Alejandro and Nilchiani, Roshanak and Verma, Dinesh},
 doi = {10.1007/s11518-016-5315-3},
 journal = {Journal of Systems Science and Systems Engineering},
 number = {5},
 pages = {549--589},
 title = {A contribution to the scientific foundations of systems engineering: {Solution} spaces and requirements},
 volume = {26},
 year = {2017}
}

@article{sexton_organizing_2020,
 author = {Sexton, Thurston and Fuge, Mark},
 doi = {10.1115/1.4045686},
 journal = {Journal of Mechanical Design},
 number = {031111},
 title = {Organizing {Tagged} {Knowledge}: {Similarity} {Measures} and {Semantic} {Fluency} in {Structure} {Mining}},
 volume = {142},
 year = {2020}
}

@article{sharma_comprehensive_2022,
 author = {Sharma, Shubhkirti and Kumar, Vijay},
 doi = {10.1007/s11831-022-09778-9},
 journal = {Archives of Computational Methods in Engineering},
 number = {7},
 pages = {5605--5633},
 title = {A {Comprehensive} {Review} on {Multi}-objective {Optimization} {Techniques}: {Past}, {Present} and {Future}},
 volume = {29},
 year = {2022}
}

@article{she_evaluating_2024,
 author = {She, Jinjuan and Belanger, Elise and Bartels, Caroline},
 doi = {10.1007/s00163-024-00434-w},
 journal = {Research in Engineering Design},
 number = {3},
 pages = {311--327},
 title = {Evaluating the effectiveness of functional decomposition in early-stage design: development and application of problem space exploration metrics},
 volume = {35},
 year = {2024}
}

@misc{shinn_reflexion_2023,
 author = {Shinn, Noah and Cassano, Federico and Berman, Edward and Gopinath, Ashwin and Narasimhan, Karthik and Yao, Shunyu},
 doi = {10.48550/arXiv.2303.11366},
 publisher = {arXiv},
 title = {Reflexion: {Language} {Agents} with {Verbal} {Reinforcement} {Learning}},
 year = {2023}
}

@misc{sumers_cognitive_2024,
 author = {Sumers, Theodore R. and Yao, Shunyu and Narasimhan, Karthik and Griffiths, Thomas L.},
 doi = {10.48550/arXiv.2309.02427},
 publisher = {arXiv},
 title = {Cognitive {Architectures} for {Language} {Agents}},
 year = {2024}
}

@article{topcu_trust_2025,
 author = {Topcu, Taylan G. and Husain, Mohammed and Ofsa, Max and Wach, Paul},
 doi = {10.1002/sys.21810},
 journal = {Systems Engineering},
 title = {Trust at {Your} {Own} {Peril}: {A} {Mixed} {Methods} {Exploration} of the {Ability} of {Large} {Language} {Models} to {Generate} {Expert}-{Like} {Systems} {Engineering} {Artifacts} and a {Characterization} of {Failure} {Modes}},
 volume = {},
 year = {2025}
}

@misc{vogel_complex_2018,
 author = {Vogel, Samuel and Rudolph, Stephan},
 doi = {10.48550/arXiv.1805.09111},
 publisher = {arXiv},
 title = {Complex {System} {Design} with {Design} {Languages}: {Method}, {Applications} and {Design} {Principles}},
 year = {2018}
}

@article{wang_survey_2024,
 author = {Wang, Lei and Ma, Chen and Feng, Xueyang and Zhang, Zeyu and Yang, Hao and Zhang, Jingsen and Chen, Zhiyuan and Tang, Jiakai and Chen, Xu and Lin, Yankai and others},
 doi = {10.1007/s11704-024-40231-1},
 journal = {Frontiers of Computer Science},
 number = {6},
 pages = {186345},
 title = {A survey on large language model based autonomous agents},
 volume = {18},
 year = {2024}
}

@inproceedings{watson_engineering_2019,
 author = {Watson, Michael D. and Mesmer, Bryan and Farrington, Phillip},
 booktitle = {Systems {Engineering} in {Context}},
 doi = {10.1007/978-3-030-00114-8_40},
 pages = {495--513},
 publisher = {Springer International Publishing},
 title = {Engineering {Elegant} {Systems}: {Postulates}, {Principles}, and {Hypotheses} of {Systems} {Engineering}},
 year = {2019}
}

@misc{wei_chain--thought_2023,
 author = {Wei, Jason and Wang, Xuezhi and Schuurmans, Dale and Bosma, Maarten and Ichter, Brian and Xia, Fei and Chi, Ed and Le, Quoc and Zhou, Denny},
 doi = {10.48550/arXiv.2201.11903},
 publisher = {arXiv},
 title = {Chain-of-{Thought} {Prompting} {Elicits} {Reasoning} in {Large} {Language} {Models}},
 year = {2023}
}

@misc{weng2023prompt,
  title   = "LLM-powered Autonomous Agents",
  author  = "Weng, Lilian",
  journal = "lilianweng.github.io",
  year    = "2023",
  month   = "Jun",
  url     = "https://lilianweng.github.io/posts/2023-06-23-agent/"
}

@article{wynn_perspectives_2017,
 author = {Wynn, David C. and Eckert, Claudia M.},
 doi = {10.1007/s00163-016-0226-3},
 journal = {Research in Engineering Design},
 number = {2},
 pages = {153--184},
 title = {Perspectives on iteration in design and development},
 volume = {28},
 year = {2017}
}

@misc{xie_human-like_2024,
 author = {Xie, Chengxing and Zou, Difan},
 doi = {10.48550/arXiv.2405.18208},
 publisher = {arXiv},
 title = {A {Human}-{Like} {Reasoning} {Framework} for {Multi}-{Phases} {Planning} {Task} with {Large} {Language} {Models}},
 year = {2024}
}

@misc{xiong_effective_2023,
 author = {Xiong, Wenhan and Liu, Jingyu and Molybog, Igor and Zhang, Hejia and Bhargava, Prajjwal and Hou, Rui and Martin, Louis and Rungta, Rashi and Sankararaman, Karthik Abinav and Oguz, Barlas and others},
 doi = {10.48550/arXiv.2309.16039},
 publisher = {arXiv},
 title = {Effective {Long}-{Context} {Scaling} of {Foundation} {Models}},
 year = {2023}
}

@misc{yao_react_2023,
 author = {Yao, Shunyu and Zhao, Jeffrey and Yu, Dian and Du, Nan and Shafran, Izhak and Narasimhan, Karthik and Cao, Yuan},
 doi = {10.48550/arXiv.2210.03629},
 publisher = {arXiv},
 title = {{ReAct}: {Synergizing} {Reasoning} and {Acting} in {Language} {Models}},
 year = {2023}
}

@misc{zhang_darwin_2025,
 author = {Zhang, Jenny and Hu, Shengran and Lu, Cong and Lange, Robert and Clune, Jeff},
 doi = {10.48550/arXiv.2505.22954},
 publisher = {arXiv},
 title = {Darwin {Godel} {Machine}: {Open}-{Ended} {Evolution} of {Self}-{Improving} {Agents}},
 year = {2025}
}

@misc{zhang_survey_2024,
 author = {Zhang, Zeyu and Bo, Xiaohe and Ma, Chen and Li, Rui and Chen, Xu and Dai, Quanyu and Zhu, Jieming and Dong, Zhenhua and Wen, Ji-Rong},
 doi = {10.48550/arXiv.2404.13501},
 publisher = {arXiv},
 title = {A {Survey} on the {Memory} {Mechanism} of {Large} {Language} {Model} based {Agents}},
 year = {2024}
}

@article{picard_concept_2025,
	title = {From concept to manufacturing: evaluating vision-language models for engineering design},
	volume = {58},
	issn = {1573-7462},
	shorttitle = {From concept to manufacturing},
	doi = {10.1007/s10462-025-11290-y},
	language = {en},
	number = {9},
	journal = {Artificial Intelligence Review},
	author = {Picard, Cyril and Edwards, Kristen M. and Doris, Anna C. and Man, Brandon and Giannone, Giorgio and Alam, Md Ferdous and Ahmed, Faez},
	month = jul,
	year = {2025},
	pages = {288},
}

@inproceedings{diniz_optimizing_2024,
	title = {Optimizing {Diffusion} to {Diffuse} {Optimal} {Designs}},
	booktitle = {{AIAA} {SCITECH} 2024 {Forum}},
	publisher = {American Institute of Aeronautics and Astronautics},
    year      = {2024},
	author = {Diniz, Cashen and Fuge, Mark},
	doi = {10.2514/6.2024-2013},
}

@inproceedings{giannone_aligning_2023,
	title = {Aligning {Optimization} {Trajectories} with {Diffusion} {Models} for {Constrained} {Design} {Generation}},
	volume = {36},
	language = {en},
	journal = {Advances in Neural Information Processing Systems},
	author = {Giannone, Giorgio and Srivastava, Akash and Winther, Ole and Ahmed, Faez},
	month = dec,
	year = {2023},
	pages = {51830--51861},
}

@article{keeler_exploring_2025,
	title = {Exploring {Human} and {Language} {Model} {Alignment} in {Perceived} {Design} {Similarity} {Using} {Ordinal} {Embeddings}},
	volume = {147},
	issn = {1050-0472},
	doi = {10.1115/1.4069129},
	number = {101402},
	journal = {Journal of Mechanical Design},
	author = {Keeler, Matthew and Fuge, Mark D. and Peng, Aoran and Miller, Scarlett},
	month = jul,
	year = {2025},
}







\appendix
\newpage
\section{Cahier des Charges: Solar-Powered Water Filtration System}
\label{sec:cahier_des_charges}

\noindent\rule{\linewidth}{0.5mm}
\begin{center}
  {\LARGE\bfseries Cahier des Charges}\\[3pt]
  {\Large Solar-Powered Water Filtration System}
\end{center}
\noindent\rule{\linewidth}{0.5mm}

\vspace{3pt}
\subsection*{A. Stakeholder Needs}
\begin{description}
  \item[SN-1] Safe drinking water in off-grid locations.
  \item[SN-2] Minimal user effort (\(\le 10\,\mathrm{min}\) routine maintenance per day).
  \item[SN-3] Affordable for target regions (\(\le \$500\) household, \(\le \$5\,000\) community).
  \item[SN-4] Environmentally responsible materials and end-of-life disposal.
  \item[SN-5] Portable (household) or easily palletized (community).
\end{description}

\vspace{3pt}
\subsection*{B. System-Level Requirements}

\begin{table}[h]
  \renewcommand{\arraystretch}{1.2}
  \centering
  \caption*{\textbf{B-1 Top-Level ``Shall'' Requirements}}
  \begin{tabular}{|p{1.6cm}|p{10.0cm}|p{2.6cm}|}
    \hline
    \textbf{ID} & \textbf{Requirement} & \textbf{Verification} \\
    \hline
    SR-01 & The system shall deliver $\ge 10\,\mathrm{L\,h^{-1}}$ potable water (25~$^\circ$C, 1~atm) from raw sources with TDS $\le 1\,000\,\mathrm{mg\,L^{-1}}$. & Test \\
    \hline
    SR-02 & The system shall achieve $\ge 4\,\log_{10}(99.99\%)$ removal of bacteria, viruses, and $1\,\mu\mathrm{m}$ micro-plastics. & Lab Analysis \\
    \hline
    SR-03 & SR-01 and SR-02 shall be met with incident solar irradiance $\ge 300\,\mathrm{W\,m^{-2}}$ (AM1.5). & Test \\
    \hline
    SR-04 & Average electrical power consumption shall be $< 50\,\mathrm{W}$ at the SR-01 flow-rate. & Analysis \\
    \hline
    SR-05 & The system shall operate $\ge 6\,\mathrm{h}$ without sunlight while maintaining SR-01 flow-rate. & Test \\
    \hline
    SR-06 & The system shall operate from $-10~^\circ\mathrm{C}$ to $50~^\circ\mathrm{C}$ and 0––95\% RH with $\le 10\%$ performance loss. & Test / Analysis \\
    \hline
    SR-07 & Dry mass shall be $< 20\,\mathrm{kg}$ (household) and $< 80\,\mathrm{kg}$ (community). & Inspection \\
    \hline
    SR-08 & $\ge 60\%$ of product mass shall be recyclable (ISO 14021); no RoHS-restricted substance above threshold. & Inspection \\
    \hline
    SR-09 & The user interface shall allow an untrained user to start/stop filtration in $\le 3$ actions and display water-quality status in $< 2\,\mathrm{s}$. & Demonstration \\
    \hline
    SR-10 & Delivered unit cost (FOB) $\le \$500$ (household) and $\le \$5\,000$ (community) at $1\,000\,\mathrm{units\,yr^{-1}}$. & Analysis \\
    \hline
  \end{tabular}
\end{table}

\vspace{6pt}
\subsection*{C. Constraints and Interfaces}
\begin{itemize}
  \item \textbf{Environmental}: Must withstand dust, rain splash (IP54 min.).
  \item \textbf{Power}: 100 \% solar with integral energy storage; external AC charger optional but not required for compliance.
  \item \textbf{Water Quality Sensors}: Output digital readings via standard UART or I\(^2\)C protocol for external logging (interface only—implementation left to design).
\end{itemize}

\vspace{6pt}
\subsection*{D. Verification Strategy}
Each SR is tagged with a primary verification method—Inspection (I), Analysis (A), Test (T), or Demonstration (D).  A detailed Requirements Verification Matrix (RVM) shall be produced during the design phase.

\vspace{6pt}
\subsection*{E. Required Design Outputs}
\begin{enumerate}
  \item \textbf{Functional Decomposition}: A hierarchy of functions and sub-functions derived from SR-01 … SR-10.
  \item \textbf{Subsystem Architecture}: Alternative mappings of functions to physical subsystems; designer/LLM is free to choose technologies.
  \item \textbf{Numerical Models}: Physics-based or empirically justified models supporting performance predictions for selected architecture(s).
  \item \textbf{Trade Study}: At least three concept variants evaluated against SR-01 … SR-10.
  \item \textbf{Verification Plan}: Test matrices, analysis methods, and pass/fail criteria traceable to each SR.
\end{enumerate}

\bigskip
\noindent\rule{\linewidth}{0.3mm}

\begin{center}
\bfseries FINALIZED
\end{center}

\section{Agent Prompts}
\label{sec:app_agent_prompts}

This appendix contains the main prompts used by each agent in the multi-agent system (MAS) and two-agent system (2AS).


\subsection{Extractor Agent Prompt}
\label{subsec:req_prompt}

\begin{code}
\begin{mintedbox}{text}
You are the Requirements Gathering Agent. Your role is to engage in a structured dialogue with the user 
to refine and finalize the technical scope of the project.

### **Your Task**
1. Extract and structure the **Cahier des Charges (Technical Scope Document)**  
2. Ask **clarifying questions** to refine missing details  
3. Ensure **functional & non-functional requirements** are well-defined  
4. Track **assumptions & open questions** for future clarification  

### **Structured Output Format**
Your output **must be valid JSON** matching this schema:
{
  "project_name": "...",
  "description": "...",
  "objectives": ["..."],
  "functional_requirements": [
    { "id": 1, "description": "..." },
    { "id": 2, "description": "..." }
  ],
  "non_functional_requirements": [
    { "id": 1, "category": "Performance", "description": "..." },
    { "id": 2, "category": "Safety", "description": "..." }
  ],
  "constraints": { "Budget": "...", "Materials": "...", "Legal": "..." },
  "assumptions": ["..."],
  "open_questions": ["..."]
}

### **Clarification Process**
- If **details are missing**, ask the user for more information.  
- If **uncertainties exist**, track them in `"open_questions"`.  
- If **finalized**, ensure `"open_questions": []` and **return 'FINALIZED'** in the response.  

**ONLY** once **fully refined**, mark the response as **FINALIZED** so the system can proceed to the planner. Do not write **FINALIZED** in your response otherwise.
If you are told to write **FINALIZED** in your response, do it.
\end{mintedbox}
\captionof{listing}{Requirements Agent prompt for structured requirements gathering.}
\end{code}


\subsection{Supervisor Agent Prompt}
\label{subsec:supervisor_prompt}

\begin{code}
\begin{mintedbox}{text}
You are the Supervisor in a multi-agent engineering-design workflow. 
The main output of this framework is a design graph that is a complete and accurate representation of the engineering system, including all subsystems, components, and their interactions.
The design graph is a mean to get to the numerical script for each subsystem/embodiement, so it can be used to simulate the system in downstream applications.
The design graph, also called Design-State Graph (DSG), must respect the specifications given by the Cahier des Charges (CDC).
You are responsible to ensure that the design graph is complete and accurate by providing feedback to the all the agents. 
Here are the agents working for you and their roles:
- Generation: Generate Design-State Graph (DSG) proposals
- Reflection: Critique the DSG proposals and provide feedback
- Ranking: Grade the DSG proposals
- Meta-Review: Select the best DSG proposal from the list of proposals

You are the boss - be assertive, directive, and clear in your instructions. Your role is to ensure the design process produces exceptional results that fully satisfy the requirements.

INPUT
• The latest Design-State Graph summary (if any)
• The original requirements (CDC): this is the only thing you get at the beginning of the process
• Meta-Review notes suggesting improvements (if any)
• Your previous instructions (if any)

TASK
Evaluate the current state and provide clear, actionable direction. You are in control of the design process: as long as the task is not satisfactory for you, it will continue to be done, and you will be revisited.
If and only if the Design-State Graph (DSG) is complete and accurate, stop the process. Otherwise, continue the process.
\end{mintedbox}
\captionof{listing}{Supervisor Agent prompt for workflow coordination.}
\end{code}


\subsection{Generator Agent Prompt (MAS)}
\label{subsec:gen_mas_prompt}

\begin{code}
\begin{mintedbox}{text}
You are the **Generation Agent** in a multi-agent systems engineering workflow.  
Your task is to produce **exactly three (3)** candidate "Design-State Graphs (DSGs)" for **<System_Name>**, each representing a different Pareto-optimal trade-off in the design space.

Each DSG must be:

1. **Complete Functional Decomposition**  
   • Break down the system **<System_Name>** into all necessary functions, sub-functions, and physical components.  
   • Show *every* subsystem or component needed to satisfy all Stakeholder Needs (SN-1 through SN-N) and System Requirements (SR-1 through SR-M).  
   • Do not leave any high-level function or lower-level component out—list everything from top-level subsystems down to atomic components that play a role in fulfilling the CDC.

2. **Accurate Traceability to the Cahier-des-Charges (CDC)**  
   • Every node in your DSG must include a `linked_reqs` field listing exactly which SRs (\eg "SR-1", "SR-2", etc.) and/or SNs it satisfies.  
   • If a particular requirement is not addressed by any node, that is not allowed—point out the missing function explicitly.  
   • The top-level design graph must show how each SR (and each SN, if applicable) is covered. If a requirement (\eg "SR-3: X must do Y") maps to multiple nodes, list them all.

3. **Complete Node Definitions Using the DSG Dataclasses**  
   For **each** `DesignNode` in a DSG, you must fill in all of the fields **completely**.

4. **No Orphan Nodes or Cycles**
• Every node must be connected—no completely isolated components unless you explicitly justify why it is a standalone leaf (\eg "Reflector" is purely decorative and has no downstream interactions).
• The graph should be acyclic, unless a feedback loop is physically and functionally justified (\eg "Control_Electronics → Actuator → Sensor → Control_Electronics" for closed-loop control).
• Each edge must represent a meaningful data/energy/material flow or interface (\eg "Pump → Filter", "Heat_Exchanger → Engine_Block").

5. **Pareto-Optimal Variations**
You must submit three distinct DSGs that differ in at least one major trade-off dimension—examples include:

    Design A (Minimum Cost): Emphasize cheapest components, minimal features, but still meet all SRs.

    Design B (Maximum Performance): Emphasize highest efficiency/throughput, advanced materials, accepting higher cost.

    Design C (Lightweight/Portable): Emphasize low mass, compactness, even if cost/performance is moderate.

    Design D (Highly Automated/Smart): Emphasize sensors/controls, digital interfaces, remote monitoring, with moderate cost.

    Design E (Maximal Recyclability/Sustainability): Emphasize recyclable materials, low environmental impact, possibly at a cost trade-off.

Each design must clearly indicate which nodes/components differ (\eg different embodiment principles or design parameters) and show how every SR and SN is still satisfied.

6. **Respect the Cahier-des-Charges (CDC) Exactly**
• Insert your actual CDC here, including all Stakeholder Needs (SN-1…SN-N) and System Requirements (SR-1…SR-M).
• Ensure the top-level design graph "<System_Name>" meets all S**.

7. **Output Format**
• For each of the designs, print exactly one JSON- or Python-serialized DesignState(...) object.
• Each DesignState must include all DesignNode entries (fully populated) and an edges list.
\end{mintedbox}
\captionof{listing}{Generation Agent prompt for creating DSG proposals in the MAS.}
\end{code}


\subsection{Reflector Agent Prompt (MAS)}
\label{subsec:reflection_mas_prompt}

\begin{code}
\begin{mintedbox}{text}
You are the Reflection agent in a multi-agent engineering design workflow.
The main output of this framework is a design graph that is a complete and accurate representation of the engineering system, including all subsystems, components, and their interactions.
The design graph is a mean to get to the numerical script for each subsystem/embodiement, so it can be used to simulate the system in downstream applications.
You are responsible to ensure that the design graph is complete and accurate and respects the supervisor instructions and the cahier des charges.

INPUT
• Current supervisor instructions for this design step.  
• The project's Cahier des Charges (CDC).  
• N Design-State Graph (DSG) proposals, each summarized in plain text.  

TASK
For each proposal (index 0 … N-1) write a concise, engineering-rigorous critique that covers:
  - Technical soundness & feasibility.  
  - Completeness w.r.t. the step objectives.  
  - Compliance with CDC requirements, objectives and constraints.  
  - Clear, actionable improvements (or explicitly state "Proposal is already optimal.").
\end{mintedbox}
\captionof{listing}{Reflection Agent prompt for critiquing DSG proposals in the MAS.}
\end{code}


\subsection{Ranker Agent Prompt}
\label{subsec:ranking_prompt}

\begin{code}
\begin{mintedbox}{text}
You are the **Ranking Agent** in a multi-agent engineering design workflow.
The main output of this framework is a design graph that is a complete and accurate representation of the engineering system, including all subsystems, components, and their interactions.
The design graph is a mean to get to the numerical script for each subsystem/embodiement, so it can be used to simulate the system in downstream applications.
You will be given a list of Design-State Graph (DSG) proposals, and your task is to grade each proposal.

Your job: give every Design-State Graph (DSG) proposal a **score 0-10, 10 being the best**
and a justification for your score.

Judge each proposal on:

1. Alignment with the current **Supervisor instructions**
2. Compliance with the **Cahier des Charges** (CDC)
3. Feedback by the **Reflection agent**
\end{mintedbox}
\captionof{listing}{Ranking Agent prompt for scoring DSG proposals.}
\end{code}


\subsection{Meta-Review Agent Prompt}
\label{subsec:meta_review_prompt}

\begin{code}
\begin{mintedbox}{text}
You are the **Meta-Review** agent in a multi-agent engineering design workflow.
The main output of this framework is a design graph that is a complete and accurate representation of the engineering system, including all subsystems, components, and their interactions.
The design graph is a mean to get to the numerical script for each subsystem/embodiement, so it can be used to simulate the system in downstream applications.
You are responsible to review the Design-State Graph (DSG) proposals, the feedback from the Reflection agent and grade (0 worst, 10 best) from the Ranking agent, consider the supervisor instructions and the cahier des charges and select the best one.
You will then inform the Superisor of your choice, the reason of your choice and the changes to the Design-State Graph (DSG) to be made, if any.

INPUT
• N design-state-graph (DSG) proposals, each with:
  - A complete DSG structure
  - Reflection feedback (technical critique and suggestions)
  - Ranking score and justification
• Supervisor instructions
• Cahier des Charges
• Current step index and iteration tracking

RULES
* Select the best Design-State Graph (DSG) proposal from the list of proposals: only one DSG is selected.
* Do **NOT** modify DSGs - only evaluate and decide
* Consider all inputs equally unless explicitly stated otherwise
* Provide clear justification for your selection
* Ensure decisions align with the current design step

OUTPUT
Return a MetaReviewOutput object with:
- selected_proposal_index: The index of your chosen solution
- detailed_summary_for_graph: Specific instructions for improving the selected solution
- decisions: List of SingleMetaDecision objects, each containing:
  - proposal_index: Index of the proposal
  - final_status: "selected", "rejected", or "needs iteration"
  - reason: Clear explanation of the decision, referencing:
    * Grade from the Ranking agent
    * Feedback from the Reflection agent
    * Current design step alignment
\end{mintedbox}
\captionof{listing}{Meta-Review Agent prompt for final proposal selection.}
\end{code}


\subsection{Coder Agent Prompt}
\label{subsec:coder_prompt}

\begin{code}
\begin{mintedbox}{text}
You are a world‐class Python coding agent with deep experience in physics‐based simulation, finite‐element methods, and multi‐physics coupling. Your output will become one node in a larger Design‐State Graph (DSG) for a complete engineering system. Every node you write must be:

  • Correct (both syntactically and physically).  
  • Fully runnable (no placeholders left behind).  
  • High‐fidelity (captures key time‐ and space‐dependent effects).  
  • Packaged as a single self‐contained Python script (no imports or file references beyond standard library, NumPy, SciPy, and pytest).

**MISSION** – Deliver ONE ready-to-run Python script (or a clearly-organised small package if multiple files really help) that serves as a high-fidelity physics / data-generation "node" in a larger pipeline.  
You may rely on robust, widely-used open-source libraries (NumPy, SciPy, matplotlib, FEniCS, PyTorch, JAX, etc.).  
Keep external dependencies minimal and justified.

--------------------------------------------------------
INPUT YOU WILL RECEIVE
--------------------------------------------------------
• node name + model name  
• governing equations (if any) and key assumptions  
• optional starting code or stubs  

--------------------------------------------------------
GENERAL EXPECTATIONS
--------------------------------------------------------
1. **Correctness & Fidelity** – code must be syntactically correct, physically meaningful, and numerically sound.  
2. **Runnable** – after `pip install -r requirements.txt` (or a short list of packages), the user can execute `python <script>` or `python -m <package>` and obtain results.  
3. **Self-Documented** – clear module docstring, inline docstrings with type hints, and a short README / usage block at the end.  
4. **CLI** – expose key parameters via `argparse` (or Typer/Click if already used).  
5. **Logging** – use Python's `logging` (or a similarly lightweight tool) with a `--verbosity` flag.  
6. **Outputs** – write results to an `./outputs` folder in at least one portable format (NumPy, CSV, VTK, HDF5, etc.) plus an optional quick-look plot.  
7. **Testing / Verification** – include a minimal pytest (or unittest) suite that checks at least one analytic or regression case.  
8. **Coupling Stub** – provide a clearly marked function that would send / receive data if this node were composited with others.  
9. **Performance** – prefer vectorised NumPy / JAX / PyTorch or sparse SciPy; avoid obvious O(N³) bottlenecks on large meshes or datasets.

--------------------------------------------------------
CONDITIONAL GUIDELINES
--------------------------------------------------------
If **PDEs are involved**  
  • Build (or import) a mesh and discretise the PDE with either  
    – a pure-Python approach (NumPy/SciPy) or  
    – a trusted library (FEniCS, Firedrake, pyMESH, etc.).  
  • Offer at least one time-integration approach that suits the physics (explicit, implicit, or adaptive).  
  • For nonlinear problems, use Newton-type iterations and log residuals.

If **only algebraic / data-driven models** (\eg, compressor maps, surrogate ML models)  
  • Focus on clean data I/O, calibration/fit routines, and prediction APIs.  
  • Provide a quick validation plot or numeric check against reference data.

--------------------------------------------------------
OUTPUT FORMAT
--------------------------------------------------------
Respond with the reasoning process and the full Python code within python tags so I can extract the code. 
If a design choice is ambiguous and reasonable defaults exist, choose one and proceed; ask the user only when absolutely necessary.
\end{mintedbox}
\captionof{listing}{Coder Agent prompt for refining numerical Python scripts.}
\end{code}


\subsection{Orchestrator Agent Prompt}
\label{subsec:orchestrator_prompt}

\begin{code}
\begin{mintedbox}{text}
You are the **Orchestrator** in a multi-agent engineering-design system.

INPUT  
• A request from another agent (Generation, Reflection, Ranking, Meta-Review, …)  
  The request always concerns a **Design-State Graph (DSG)** proposal or its critique.

TASK  
Break the request into at most **three** concrete Worker tasks that involve
  • Web or ArXiv searches  
  • Light calculations or code snippets (if explicitly asked)  

For **each** task return:  
- `"topic"` : a 1-line title  
- `"description"` : what to search / calculate and **why** it helps the requesting agent  

If no external work is needed, set `"tasks": []` and put a short explanation in `"response"`.

Be precise; avoid vague or duplicate tasks.
\end{mintedbox}
\captionof{listing}{Orchestrator Agent prompt for task decomposition.}
\end{code}


\subsection{Worker Agent Prompt}
\label{subsec:worker_prompt}

\begin{code}
\begin{mintedbox}{text}
You are a **Worker Agent** in the engineering-design workflow.

INPUT  
• A single task from the Orchestrator (Web/ArXiv search or lightweight calculation).  
• Each task supports analysis or improvement of a **Design-State Graph (DSG)**.

TOOLS  
- **Web Search** (find standards, data, component specs, etc.)  
- **ArXiv Search** (find peer-reviewed methods or equations)  
- (Optional) lightweight Python snippets if explicit.

OUTPUT  (structured, concise)  
1. **Findings** – key facts, equations, or data (cite sources/links).  
2. **Design insight** – how these findings help refine or validate the DSG.  

If information is insufficient, state limitations and suggest next steps.
\end{mintedbox}
\captionof{listing}{Worker Agent prompt for executing research tasks.}
\end{code}


\subsection{Generator Agent Prompt (2AS)}
\label{subsec:gen_2as_prompt}

\begin{code}
\begin{mintedbox}{text}
You are the **Generation Agent** in a two-agent systems engineering workflow. It is you and the Reflection agent that are working together to design an engineering system.
If iteration is needed, you will be informed by the Reflection agent.
Your task is to produce **exactly three (3)** candidate "Design-State Graphs (DSGs)" for **<System_Name>**, each representing a different Pareto-optimal trade-off in the design space.

Each DSG must be:

1. **Complete Functional Decomposition**  
   • Break down the system **<System_Name>** into all necessary functions, sub-functions, and physical components.  
   • Show *every* subsystem or component needed to satisfy all Stakeholder Needs (SN-1 through SN-N) and System Requirements (SR-1 through SR-M).  
   • Do not leave any high-level function or lower-level component out—list everything from top-level subsystems down to atomic components that play a role in fulfilling the CDC.

2. **Accurate Traceability to the Cahier-des-Charges (CDC)**  
   • Every node in your DSG must include a `linked_reqs` field listing exactly which SRs (\eg "SR-1", "SR-2", etc.) and/or SNs it satisfies.  
   • If a particular requirement is not addressed by any node, that is not allowed—point out the missing function explicitly.  
   • The top-level design graph must show how each SR (and each SN, if applicable) is covered. If a requirement (\eg "SR-3: X must do Y") maps to multiple nodes, list them all.

3. **Complete Node Definitions Using the DSG Dataclasses**  
   For **each** `DesignNode` in a DSG, you must fill in all of the fields **completely**.

4. **No Orphan Nodes or Cycles**
• Every node must be connected—no completely isolated components unless you explicitly justify why it is a standalone leaf (\eg "Reflector" is purely decorative and has no downstream interactions).
• The graph should be acyclic, unless a feedback loop is physically and functionally justified (\eg "Control_Electronics → Actuator → Sensor → Control_Electronics" for closed-loop control).
• Each edge must represent a meaningful data/energy/material flow or interface (\eg "Pump → Filter", "Heat_Exchanger → Engine_Block").

5. **Pareto-Optimal Variations**
You must submit three distinct DSGs that differ in at least one major trade-off dimension—examples include:

    Design A (Minimum Cost): Emphasize cheapest components, minimal features, but still meet all SRs.

    Design B (Maximum Performance): Emphasize highest efficiency/throughput, advanced materials, accepting higher cost.

    Design C (Lightweight/Portable): Emphasize low mass, compactness, even if cost/performance is moderate.

    Design D (Highly Automated/Smart): Emphasize sensors/controls, digital interfaces, remote monitoring, with moderate cost.

    Design E (Maximal Recyclability/Sustainability): Emphasize recyclable materials, low environmental impact, possibly at a cost trade-off.

Each design must clearly indicate which nodes/components differ (\eg different embodiment principles or design parameters) and show how every SR and SN is still satisfied.

6. **Respect the Cahier-des-Charges (CDC) Exactly**
• Insert your actual CDC here, including all Stakeholder Needs (SN-1…SN-N) and System Requirements (SR-1…SR-M).
• Ensure the top-level design graph "<System_Name>" meets all S**.

7. **Output Format**
• For each of the designs, print exactly one JSON- or Python-serialized DesignState(...) object.
• Each DesignState must include all DesignNode entries (fully populated) and an edges list.
\end{mintedbox}
\captionof{listing}{Generation Agent prompt for the two-agent system (2AS).}
\end{code}


\subsection{Reflector Agent Prompt (2AS)}
\label{subsec:reflection_2as_prompt}

\begin{code}
\begin{mintedbox}{text}
You are the Reflection agent in a two-agent engineering design workflow. Yourself and the Generation agent are working together to design an engineering system.
The main output of this framework is a design graph that is a complete and accurate representation of the engineering system, including all subsystems, components, and their interactions.
The design graph is a mean to get to the numerical script for each subsystem/embodiement, so it can be used to simulate the system in downstream applications.
You are responsible to ensure that the design graph is complete and accurate and respects the cahier des charges.

INPUT
• Current supervisor instructions for this design step.  
• The project's Cahier des Charges (CDC).  
• N Design-State Graph (DSG) proposals, each summarized in plain text.  

TASK
For each proposal (index 0 … N-1) write a concise, engineering-rigorous critique that covers:
  - Technical soundness & feasibility.  
  - Completeness w.r.t. the step objectives.  
  - Compliance with CDC requirements, objectives and constraints.  
  - Clear, actionable improvements (or explicitly state "Proposal is already optimal.").
Then select the best proposal and give it a selected status. You must select ONE and ONLY ONE proposal.
Give a rejected status to all other proposals.
A proposal may be relatively the best compared to the others, but still needs more iteration. If this the case, explain how it can be improved.
Give a reason for your decision.

Terminate the workflow when the DSG is complete and accurate and respects the cahier des charges.

Your feedback guides the Generation agent to improve the selected design graph.
\end{mintedbox}
\captionof{listing}{Reflection Agent prompt for the two-agent system (2AS).}
\end{code}


\section{Extracted Node Code Listings}
\label{sec:app_code_listings}


\vspace{-2mm}
\subsection{MAS DSG – Solar Power Generation Node}
\label{subsec:mas_solar_python_code}

\noindent\textbf{Embodiment:} Flexible thin‐film solar panels (15\% eff., 100 W), lightweight.  

\begin{code}
\begin{mintedbox}{python}
#!/usr/bin/env python3
"""
solar_power.py

Calculate energy produced by a flexible thin‐film solar panel:
    P = eta * S * t
Includes CLI, logging, output saving.
"""

import argparse
import logging
import numpy as np
import matplotlib.pyplot as plt
import os

# Set up logging
logging.basicConfig(
    level=logging.INFO,
    format='%(asctime)s - %(levelname)s - %(message)s'
)
logger = logging.getLogger(__name__)

def calculate_power(efficiency: float, solar_irradiance: float = 1000.0, time: float = 1.0) -> float:
    """
    Calculate the power generated by a solar panel.
    P = eta * S * t  (Wh)
    """
    logger.info(f"Calculating power with efficiency {efficiency}, irradiance {solar_irradiance}, time {time}")
    if efficiency <= 0 or efficiency >= 1:
        raise ValueError("Efficiency must be between 0 and 1")
    if solar_irradiance <= 0 or time <= 0:
        raise ValueError("Solar irradiance and time must be positive")
    power = efficiency * solar_irradiance * time
    logger.info(f"Calculated power: {power} Wh")
    return power

def setup_args():
    parser = argparse.ArgumentParser(description='Calculate solar power generation')
    parser.add_argument('--efficiency', type=float, default=0.2,
                        help='Panel efficiency (0–1)')
    parser.add_argument('--time', type=float, default=1.0,
                        help='Operation time in hours')
    parser.add_argument('--verbosity', type=int, default=2,
                        help='Logging verbosity level (0–4)')
    return parser.parse_args()

def save_results(power: float, output_dir: str = './outputs'):
    os.makedirs(output_dir, exist_ok=True)
    np.save(os.path.join(output_dir, 'power_generation.npy'), power)
    np.savetxt(os.path.join(output_dir, 'power_generation.csv'), [power], header='Power(Wh)')
    plt.figure(figsize=(8,6))
    plt.title('Solar Power Generation')
    plt.text(0.1,0.5,f'Power: {power} Wh',ha='left',va='center')
    plt.axis('off')
    plt.savefig(os.path.join(output_dir, 'power_generation.png'))
    plt.close()
    logger.info(f"Results saved to {output_dir}")

def main():
    args = setup_args()
    level = {0:logging.CRITICAL,1:logging.ERROR,2:logging.INFO}.get(args.verbosity,logging.DEBUG)
    logger.setLevel(level)

    try:
        p = calculate_power(args.efficiency, time=args.time)
        save_results(p)
    except Exception as e:
        logger.error(f"Error: {e}")
        raise

if __name__ == "__main__":
    main()
\end{mintedbox}
\captionof{listing}{Full Python code for the Solar Power Generation node (MAS DSG).}
\end{code}

\vspace{-2mm}
\subsection{MAS DSG – Water Filtration Node}
\label{subsec:mas_filtration_python_code}

\noindent\textbf{Embodiment:} Ceramic microfiltration (0.2 \ensuremath{\mu m} pore).  

\begin{code}
\begin{mintedbox}{python}
#!/usr/bin/env python3
"""
flow_rate.py

Compute flow rate through a ceramic water filter:
    Q = (filter_area * permeability) / viscosity
"""

import argparse
import logging
import os
import numpy as np
import matplotlib.pyplot as plt

# Logging setup
logging.basicConfig(level=logging.INFO, format='%(asctime)s - %(message)s')
logger = logging.getLogger(__name__)

def calculate_flow_rate(filter_area: float, permeability: float, viscosity: float) -> float:
    """
    Q = (A * kappa) / mu
    """
    if filter_area <= 0 or permeability <= 0 or viscosity <= 0:
        raise ValueError("All inputs must be positive")
    return (filter_area * permeability) / viscosity

def main():
    parser = argparse.ArgumentParser(description="Ceramic Filter Flow Rate")
    parser.add_argument("--area", type=float, required=True, help="Filter area (m^2)")
    parser.add_argument("--perm", type=float, required=True, help="Permeability")
    parser.add_argument("--visc", type=float, required=True, help="Viscosity (Pa·s)")
    args = parser.parse_args()

    Q = calculate_flow_rate(args.area, args.perm, args.visc)
    logger.info(f"Flow rate: {Q:.6f} m^3/s")
    os.makedirs("outputs", exist_ok=True)
    np.savetxt("outputs/flow_rate.csv", [Q], header="Q (m3/s)")

if __name__ == "__main__":
    main()
\end{mintedbox}
\captionof{listing}{Full Python code for the Water Filtration node (MAS DSG).}
\end{code}

\vspace{-2mm}
\subsection{MAS DSG – Water Pump Node}
\label{subsec:mas_pump_python_code}

\noindent\textbf{Embodiment:} Low‐weight DC diaphragm pump (10 L/h, 20 W).  

\begin{code}
\begin{mintedbox}{python}
#!/usr/bin/env python3
"""
pump_performance.py

Model pump performance:
    Q = k * \omega
Includes CLI, logging, output.
"""

import argparse
import logging
import matplotlib.pyplot as plt
import os

logging.basicConfig(level=logging.INFO, format='%(asctime)s - %(message)s')
logger = logging.getLogger(__name__)

def calculate_performance(k: float, omega: float) -> float:
    """
    Q = k * omega
    """
    if k <= 0 or omega <= 0:
        raise ValueError("k and omega must be positive")
    return k * omega

def main():
    p = argparse.ArgumentParser(description="Pump Performance")
    p.add_argument("--k", type=float, required=True, help="Pump constant")
    p.add_argument("--omega", type=float, required=True, help="Angular velocity (rad/s)")
    args = p.parse_args()

    Q = calculate_performance(args.k, args.omega)
    logger.info(f"Pump flow rate: {Q:.3f} m^3/s")
    os.makedirs("outputs", exist_ok=True)
    with open("outputs/pump_performance.txt","w") as f:
        f.write(str(Q))

if __name__ == "__main__":
    main()
\end{mintedbox}
\captionof{listing}{Full Python code for the Water Pump node (MAS DSG).}
\end{code}

\vspace{-2mm}
\subsection{MAS DSG – Energy Storage Node}
\label{subsec:mas_storage_python_code}

\noindent\textbf{Embodiment:} Compact lithium‐ion battery (12 V, 10 Ah).  

\begin{code}
\begin{mintedbox}{python}
#!/usr/bin/env python3
"""
energy_storage.py

Compute stored energy:
    E = V * C * \eta
"""

import argparse
import logging

logging.basicConfig(level=logging.INFO)
logger = logging.getLogger(__name__)

def calculate_energy(voltage: float, capacity: float, efficiency: float = 1.0) -> float:
    """
    E = V * C * eta
    """
    if voltage <= 0 or capacity <= 0:
        raise ValueError("Voltage and capacity must be positive")
    return voltage * capacity * efficiency

if __name__ == "__main__":
    import argparse
    p = argparse.ArgumentParser(description="Energy Storage Calc")
    p.add_argument("voltage", type=float, help="Voltage (V)")
    p.add_argument("capacity", type=float, help="Capacity (Ah)")
    p.add_argument("--eff", type=float, default=1.0, help="Efficiency (0–1)")
    args = p.parse_args()

    E = calculate_energy(args.voltage, args.capacity, args.eff)
    logger.info(f"Stored energy: {E:.2f} Wh")
\end{mintedbox}
\captionof{listing}{Full Python code for the Energy Storage node (MAS DSG).}
\end{code}

\vspace{-2mm}
\subsection{MAS DSG – Control Unit Node}
\label{subsec:mas_control_python_code}

\noindent\textbf{Embodiment:} Low‐power microcontroller.  

\begin{code}
\begin{mintedbox}{python}
#!/usr/bin/env python3
"""
power_consumption.py

Compute control‐unit power:
    P = I * V
"""

import argparse
import logging

logging.basicConfig(level=logging.INFO)
logger = logging.getLogger(__name__)

def calculate_power(current: float, voltage: float) -> float:
    """
    P = I * V
    """
    if current < 0 or voltage < 0:
        raise ValueError("Current and voltage must be non-negative")
    return current * voltage

if __name__ == "__main__":
    import argparse
    p = argparse.ArgumentParser(description="Power Consumption")
    p.add_argument("current", type=float, help="Current (A)")
    p.add_argument("voltage", type=float, help="Voltage (V)")
    args = p.parse_args()

    P = calculate_power(args.current, args.voltage)
    logger.info(f"Consumption: {P:.2f} W")
\end{mintedbox}
\captionof{listing}{Full Python code for the Control Unit node (MAS DSG).}
\end{code}

\vspace{-2mm}
\subsection{MAS DSG – Water Quality Sensors Node}
\label{subsec:mas_sensors_python_code}

\noindent\textbf{Embodiment:} pH/turbidity/TDS sensors.  

\begin{code}
\begin{mintedbox}{python}
#!/usr/bin/env python3
"""
sensor_response.py

Compute sensor response:
    R = delay + processing
"""

import argparse
import logging

logging.basicConfig(level=logging.INFO)
logger = logging.getLogger(__name__)

def sensor_response(delay: float, processing: float) -> float:
    """
    R = delay + processing
    """
    if delay < 0 or processing < 0:
        raise ValueError("Inputs must be non-negative")
    return delay + processing

if __name__ == "__main__":
    import argparse
    p = argparse.ArgumentParser(description="Sensor Response")
    p.add_argument("delay", type=float, help="Sensor delay (s)")
    p.add_argument("processing", type=float, help="Processing time (s)")
    args = p.parse_args()

    R = sensor_response(args.delay, args.processing)
    logger.info(f"Response time: {R:.3f} s")
\end{mintedbox}
\captionof{listing}{Full Python code for the Water Quality Sensors node (MAS DSG).}
\end{code}

\vspace{-2mm}
\subsection{MAS DSG – User Interface Node}
\label{subsec:mas_ui_python_code}

\noindent\textbf{Embodiment:} LCD+buttons.  

\begin{code}
\begin{mintedbox}{python}
#!/usr/bin/env python3
"""
user_interaction.py

Compute UI response:
    R = T + E
"""

import argparse
import logging

logging.basicConfig(level=logging.INFO)
logger = logging.getLogger(__name__)

def calculate_response(time_s: float, error: float) -> float:
    """
    R = T + E
    """
    if time_s < 0 or error < 0:
        raise ValueError("Inputs must be non-negative")
    return time_s + error

if __name__ == "__main__":
    import argparse
    p = argparse.ArgumentParser(description="UI Response Time")
    p.add_argument("time", type=float, help="Base response time (s)")
    p.add_argument("error", type=float, help="Error (s)")
    args = p.parse_args()

    R = calculate_response(args.time, args.error)
    logger.info(f"Total response: {R:.3f} s")
\end{mintedbox}
\captionof{listing}{Full Python code for the User Interface node (MAS DSG).}
\end{code}

\vspace{-2mm}
\subsection{MAS DSG – Housing Node}
\label{subsec:mas_housing_python_code}

\noindent\textbf{Embodiment:} Polypropylene enclosure.  

\begin{code}
\begin{mintedbox}{python}
#!/usr/bin/env python3
"""
structural_integrity.py

Compute structural stress:
    S = F / A
"""

import argparse
import logging

logging.basicConfig(level=logging.INFO)
logger = logging.getLogger(__name__)

def calculate_stress(force: float, area: float) -> float:
    """
    S = F/A
    """
    if area <= 0:
        raise ValueError("Area must be positive")
    return force / area

if __name__ == "__main__":
    import argparse
    p = argparse.ArgumentParser(description="Structural Stress")
    p.add_argument("force", type=float, help="Force (N)")
    p.add_argument("area", type=float, help="Area (m^2)")
    args = p.parse_args()

    S = calculate_stress(args.force, args.area)
    logger.info(f"Stress: {S:.2f} Pa")
\end{mintedbox}
\captionof{listing}{Full Python code for the Housing node (MAS DSG).}
\end{code}


\vspace{-2mm}
\subsection{2AS DSG – Solar Power Subsystem (Energy Budget)}
\label{subsec:2as_energy_python_code}

\noindent\textbf{Embodiment:} Rigid PV + lead‐acid battery.  

\begin{code}
\begin{mintedbox}{python}
# energy_budget.py

"""
Energy budget for rigid PV + lead‐acid battery:
    E = irradiance * 0.15 * time
"""

import argparse
import logging

logging.basicConfig(level=logging.INFO)
logger = logging.getLogger(__name__)

def energy_budget(irradiance: float, time_h: float) -> float:
    if irradiance <= 0 or time_h <= 0:
        raise ValueError("Inputs must be positive")
    return irradiance * 0.15 * time_h

if __name__ == "__main__":
    p = argparse.ArgumentParser(description="Energy Budget Calculator")
    p.add_argument("irradiance", type=float, help="Solar irradiance (W/m²)")
    p.add_argument("time",       type=float, help="Time in hours")
    args = p.parse_args()

    E = energy_budget(args.irradiance, args.time)
    logger.info(f"Total energy: {E:.2f} Wh")
\end{mintedbox}
\captionof{listing}{Full Python code for the Solar Power Subsystem node (2AS DSG).}
\end{code}

\vspace{-2mm}
\subsection{2AS DSG – Filtration Subsystem (Flow Rate)}
\label{subsec:2as_flow_python_code}

\noindent\textbf{Embodiment:} Ceramic water filter.  

\begin{code}
\begin{mintedbox}{python}
# flow_rate.py

"""
Flow rate through ceramic filter:
    Q = (area * permeability) / viscosity
"""

import argparse
import logging

logging.basicConfig(level=logging.INFO)
logger = logging.getLogger(__name__)

def flow_rate(area: float, permeability: float, viscosity: float) -> float:
    if area <= 0 or permeability <= 0 or viscosity <= 0:
        raise ValueError("All inputs must be positive")
    return (area * permeability) / viscosity

if __name__ == "__main__":
    p = argparse.ArgumentParser(description="Filtration Flow Rate")
    p.add_argument("area",        type=float, help="Filter area (m²)")
    p.add_argument("permeability",type=float, help="Permeability")
    p.add_argument("viscosity",   type=float, help="Viscosity (Pa·s)")
    args = p.parse_args()

    Q = flow_rate(args.area, args.permeability, args.viscosity)
    logger.info(f"Flow rate: {Q:.6f} m3/s")
\end{mintedbox}
\captionof{listing}{Full Python code for the Filtration Subsystem node (2AS DSG).}
\end{code}

\vspace{-2mm}
\subsection{2AS DSG – Water Quality Monitoring Subsystem}
\label{subsec:2as_monitor_python_code}

\noindent\textbf{Embodiment:} Digital pH/turbidity/TDS sensors.  

\begin{code}
\begin{mintedbox}{python}
# sensor_response.py

"""
Compute water‐quality sensor response:
    R = delay + processing
"""

import argparse
import logging

logging.basicConfig(level=logging.INFO)
logger = logging.getLogger(__name__)

def sensor_response(delay: float, processing: float) -> float:
    if delay < 0 or processing < 0:
        raise ValueError("Inputs must be non‐negative")
    return delay + processing

if __name__ == "__main__":
    p = argparse.ArgumentParser(description="Sensor Response")
    p.add_argument("delay",      type=float, help="Sensor delay (s)")
    p.add_argument("processing", type=float, help="Processing time (s)")
    args = p.parse_args()

    R = sensor_response(args.delay, args.processing)
    logger.info(f"Response time: {R:.3f} s")
\end{mintedbox}
\captionof{listing}{Full Python code for the Water Quality Monitoring Subsystem node (2AS DSG).}
\end{code}

\vspace{-2mm}
\subsection{2AS DSG – Power Management Subsystem}
\label{subsec:2as_power_python_code}

\noindent\textbf{Embodiment:} Basic charge controller + BMS.  

\begin{code}
\begin{mintedbox}{python}
# battery_level.py

"""
Compute battery level over time:
    Level = initial_charge - discharge_rate * time
"""

import argparse
import logging

logging.basicConfig(level=logging.INFO)
logger = logging.getLogger(__name__)

def battery_level(initial_charge: float, discharge_rate: float, time_h: float) -> float:
    if initial_charge < 0 or discharge_rate < 0 or time_h < 0:
        raise ValueError("Inputs must be non‐negative")
    return initial_charge - discharge_rate * time_h

if __name__ == "__main__":
    p = argparse.ArgumentParser(description="Battery Level Calculator")
    p.add_argument("initial_charge", type=float, help="Initial charge (Wh)")
    p.add_argument("discharge_rate", type=float, help="Discharge rate (W/hour)")
    p.add_argument("time",           type=float, help="Time (h)")
    args = p.parse_args()

    L = battery_level(args.initial_charge, args.discharge_rate, args.time)
    logger.info(f"Battery level: {L:.2f} Wh")
\end{mintedbox}
\captionof{listing}{Full Python code for the Power Management Subsystem node (2AS DSG).}
\end{code}

\end{document}